\documentclass[10pt,journal,compsoc]{IEEEtran}
\ifCLASSOPTIONcompsoc
\else
\fi

\usepackage[utf8]{inputenc} 
\usepackage[T1]{fontenc}    
\usepackage{hyperref}       
\usepackage{url}            
\usepackage{booktabs}       
\usepackage{amsfonts}       
\usepackage{nicefrac}       
\usepackage{microtype}      
\usepackage{bm}
\usepackage{enumitem}
\usepackage{floatrow}
\usepackage{algorithm}
\usepackage{algorithmic}
\usepackage{subfigure}
\usepackage{graphicx}
\usepackage{epstopdf}
\usepackage{multirow}
\usepackage{amsmath}
\usepackage{amssymb}
\usepackage{caption}
\usepackage[american]{babel}

\newcommand{\argmax}{\mathop{\arg\max}}

\newcommand{\x}{{\bf x}}

\newcommand{\p}{{\bf p}}

\newcommand{\y}{{\bf y}}

\newcommand{\ie}{{\it i.e.}}
\newcommand{\eg}{{\it e.g.}}

\ifCLASSINFOpdf
\else
\fi

\begin{document}
%
\title{Revisiting Unsupervised Meta-Learning \\via the Characteristics of Few-Shot Tasks}
%
%
%
%

\author{Han-Jia Ye,
	    Lu Han,
        De-Chuan Zhan
\IEEEcompsocitemizethanks{
\IEEEcompsocthanksitem H.-J. Ye, L. Han, and D.-C. Zhan are with State Key Laboratory for Novel Software Technology,
       Nanjing University,  Nanjing, 210023, China.
       \protect\\
E-mail: \{yehj, hanlu, zhandc\}@lamda.nju.edu.cn}
}

%
%

\markboth{Journal of \LaTeX\ Class Files,~Vol.~Xx, No.~X, Xxxx~20Xx}%
{Ye \MakeLowercase{\textit{et al.}}: Revisiting Unsupervised Meta-Learning via the Characteristics of Few-Shot Tasks}
%



\IEEEtitleabstractindextext{%
\begin{abstract}
Meta-learning has become a practical approach towards few-shot image classification, where ``a strategy to learn a classifier'' is meta-learned on labeled base classes and can be applied to tasks with novel classes. 
We remove the requirement of base class labels and learn generalizable embeddings via {\em Unsupervised} Meta-Learning (UML).
Specifically, episodes of tasks are constructed with data augmentations from unlabeled base classes during meta-training, and we apply embedding-based classifiers to novel tasks with labeled few-shot examples during meta-test.
We observe two elements play important roles in UML, \ie, the way to sample tasks and measure similarities between instances. Thus we obtain a strong baseline with two simple modifications --- a sufficient sampling strategy constructing multiple tasks per episode efficiently together with a semi-normalized similarity.
We then take advantage of the characteristics of tasks from two directions to get further improvements.
First, synthesized confusing instances are incorporated to help extract more discriminative embeddings. 
Second, we utilize an additional task-specific embedding transformation as an auxiliary component during meta-training to promote the generalization ability of the pre-adapted embeddings.
Experiments on few-shot learning benchmarks verify that our approaches outperform previous UML methods and achieve comparable or even {\em better} performance than its supervised variants.
\end{abstract}

}

\maketitle


%
\IEEEpeerreviewmaketitle

\IEEEraisesectionheading{\section{Introduction}}\label{sec:intro}
\IEEEPARstart{T}{he} Few-Shot Learning~(FSL) ability~\cite{LakeSGT11One,lake2015human}, \ie, training a model with limited data, is essential in various fields, \eg, visual recognition~\cite{koch2015siamese,VinyalsBLKW16Matching,Ye2021Heterogeneous} and object detection~\cite{Fan2020Few,Wu2020Meta,Yang2020Context}.\footnote{The ``shot'' means a training example per class.}
A large number of labeled classes are collected at first. By learning on these {\em base} classes, our goal is to enable the FSL over non-overlapping {\em novel} classes --- a model trained on a novel few-shot support set should recognize new instances of those novel classes. Generalizable component across base and novel classes such as embedding is the key to FSL.

Meta-learning has become one popular approach for FSL, where a meta-model encodes a generalizable ``learning strategy'' to train a classifier given a few-shot support set~\cite{VinyalsBLKW16Matching,FinnAL17Model}.
In detail, episodes of (pseudo) tasks from base classes are sampled to mimic the target FSL scenario on novel classes, where each few-shot support set of a task is associated with a query set sharing the same set of classes. We optimize the meta-model to make the ``learning strategy'' facilitate the query set classification conditioned on the support set. 
For example, we can implement the meta-model via embedding function (the feature extractor). Then the ``learning strategy'' becomes an embedding-based classifier, which predicts a query instance through its neighbors in the support set. By minimizing classification losses over sampled tasks, the meta-model is expected to extend the effectiveness of its ``learning strategy'' to few-shot support set with novel classes.
Besides embeddings~\cite{VinyalsBLKW16Matching,SnellSZ17Prototypical,Lee2019Meta}, we can also use optimizers~\cite{FinnAL17Model,Sachin2017,Nichol2018On} and image generators~\cite{Wang2018Low} in meta-learning.

\begin{figure}
	\centering
	\includegraphics[width=1\linewidth]{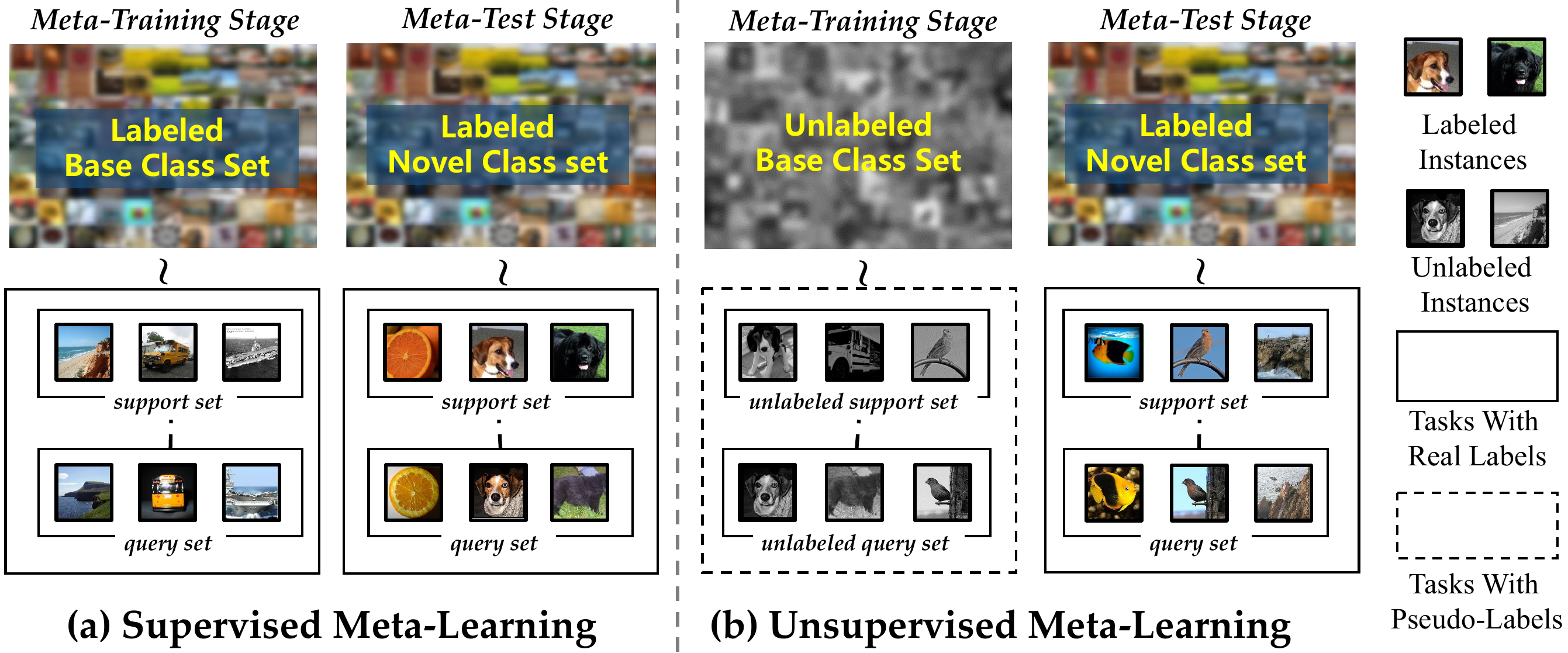}
	\caption{Supervised meta-learning (a) and Unsupervised Meta-Learning~(UML) (b) for few-shot image classification. 
	The learned meta-model should be applied to novel classes few-shot tasks after meta-training on base classes.
    Although given {\em unlabeled} base classes, UML still facilitates the few-shot model construction on novel classes.}\label{fig:teaser}
\end{figure}

We illustrate (supervised) meta-learning in Fig.~\ref{fig:teaser} (a). 
Meta-learning reduces the number of labels during deployment on novel class tasks but requires a large annotated base class set.
Labels in the base class set are essential to sample synthetic classification tasks during meta-training, and the label size is highly related to the meta-level generalization ability~\cite{Chao2020Revisiting}. In practice, we usually take advantage of those labels and pre-train a classifier discerning base classes to initialize the meta-model~\cite{Qiao2017Few,YeHZS2018Learning,Tian2020Rethinking}.
Although labeled base classes help meta-learning, we should not overlook the expense of labeling them or assume that we will always have plenty of labels in meta-training.
In the long run, meta-learning with less base class labeling is critical for FSL.

We investigate whether we can get a few-shot classifier, implemented based on embedding, successfully {\em via meta-training in an unsupervised manner}. The notion of Unsupervised Meta-Learning~(UML) is shown in Fig.~\ref{fig:teaser} (b). 
We save the labeling cost of the base classes in the holistic few-shot learning process while maintaining the generalizable ability of the meta-model. In other words, we only need a labeled few-shot support set in the meta-test phase.

Our main idea for UML follows the Supervised Meta-Learning~(SML) but defines the ``class'' based on data augmentations~\cite{dosovitskiy2014discriminative,wu2018unsupervised,khodadadeh2019unsupervised} --- any two instances have different pseudo-class labels, and an instance is only similar to the copies augmented from itself.
Then by sampling base class tasks following the pseudo-classes, the embedding could be meta-learned in the same way as SML. 
Empirical observations indicate that data augmentation makes reusing SML techniques in UML with minimal changes, but usually could not provide competitive results~\cite{antoniou2019assume,khodadadeh2019unsupervised,qin2020unsupervised}  when compared with other embedding learning paradigms like self-supervised learning~\cite{he2020momentum,chen2020simple}.

We take a closer look at UML and analyze the key factors during meta-training. We observe that the strategy to sample (pseudo) tasks from the base class set and the way to measure the similarity between instances in a task play essential roles.
To take full advantage of instances in a mini-batch, we propose Sufficient Episodic Sampling~(SES), which efficiently re-samples multiple few-shot tasks from the same mini-batch. SES produces stable gradients without additional cost to extract features.
Various similarity metrics are investigated in FSL~\cite{VinyalsBLKW16Matching,SnellSZ17Prototypical} , but they need careful calibrations via temperature tunning~\cite{Gidaris2018Dynamic,YeHZS2018Learning}. 
We propose a novel Semi-Normalized Similarity~(SNS) where only one instance in the input pair is normalized before computing their inner product. This similarity equals cosine similarity with a self-learned temperature, which softens the logits adaptively and shapes a discriminative embedding space.

The two simple modifications construct a strong UML baseline, which shows performance superiority over both UML and self-supervised learning methods on FSL benchmarks. 
Moreover, we delve into the {\em characteristics of tasks} and propose to further improve the discriminative as well as generalization ability of the learned UML embeddings.

First, we amplify the contrastiveness of instances in a task by synthesizing a more difficult task for each query instance. Specifically, we augment the support set dynamically through Hard Supports Mix (HMS). HMS constructs confusing instances by mixing up the query instances with the nearest ones belonging to different pseudo-classes. Differentiating over those confusing instances leads to more discriminative embeddings that generalize better to {\em lower-shot} (\eg, 1-shot) tasks.
On the other hand, since tasks from base and novels classes are constructed based on pseudo-labels and real semantic labels, respectively, their distribution gap makes the generalization of the meta-model difficult. We mitigate the negative effects with an auxiliary Task-Specific Projection Head (TSP-Head) in meta-training.
We decompose the adapted embedding function into generalizable and specific parts, where the auxiliary transformation handles the specific property of pseudo-labeled tasks and it makes the vanilla embedding generalize better, especially with {\em relatively higher-shot} (\eg, 20-shot). In other words, during meta-training, we construct the embedding-based classifier using the adapted embeddings, but in deployment, we utilize the pre-adapted embedding before the TSP-Head.

Experiments on few-shot classification and cross-domain benchmarks demonstrate that our UML methods outperform other unsupervised, self-supervised, or even SML methods. Our main contributions could be summarized as follows:
\begin{itemize}
	\item We analyze the factors to meta-train a UML method and propose SES and SNS as two key ingredients towards a strong UML baseline.
	\item We propose HMS and TSP-Head to further utilize the characteristic of tasks from different aspects, which additionally improve either lower or higher shots scenarios.
	\item Our UML methods outperform existing ones by a large margin with unlabeled base classes and even get better results when compared with supervised counterparts.
\end{itemize}
The remaining parts start with related work and preliminaries. 
After a close study of key components in UML, we propose our strong UML baseline. Then we demonstrate the importance of considering the characteristic of tasks for further improvements. Last are experiments and conclusions.

\section{Related Work}
\label{sec:related}
{\bf Meta-learning for Few-Shot Learning~(FSL).} Training a high-quality visual recognition system usually requires an ample number of annotated training set with many shots~\cite{RussakovskyDSKS15ImageNet,Krizhevsky2017ImageNet}, and a few-shot training set makes the model prone to overfitting.
FSL aims to enable the classification on a few-shot support set with novel classes given labeled data-rich base classes~\cite{Fei-FeiFP06One,FinnAL17Model,LiZCL17Meta,Nichol2018On,Lee2018Gradient}.
Meta-learning has become an effective tool for FSL, which generalizes a ``learning strategy'' from base to novel classes. 
One main thread of meta-learning considers transferable embeddings~\cite{SnellSZ17Prototypical,Oreshkin2018TADAM,Rusu2018Meta,Scott2018Adapted,Lee2019Meta,YeHZS2018Learning}, so that with the help of a non-parametric nearest neighbor classifier, novel class instances could be recognized given a few labeled examples. 
Other meta-learning methods explore the meta-model with initialization~\cite{FinnAL17Model,Nichol2018On,Ye2021How}, optimization policies~\cite{Sachin2017,Rusu2018Meta}, image generator~\cite{Wang2018Low}, and the mapping from data to classifier~\cite{WangH16Learning,WangRH17Learning}.
\cite{Yue2020Interventional} analyzes FSL in a causal view.
FSL has achieved promising results in various domains~\cite{TriantafillouZU17Few,Dong2018Domain,Kang2018Transferable,Lifchitz2019Dense,Pang2021Few}. Empirical studies of FSL are in~\cite{chen2019closer,triantafillou2019meta,Chao2020Revisiting,Ye2021Few}.

\noindent{\bf Meta-learning with unlabeled data.}
Despite the success of FSL on novel classes in deployment, meta-learning requires plenty of base class labels during meta-training. 
One intuitive way is to utilize unlabeled data during episodic learning. 
In semi-supervised meta-learning, there is a pool set containing unlabeled instances even from distractor classes associated with the support set~\cite{Ren2018Meta,Ye2021TACO,Yu2020TransMatch,Zhu2022Label}. Transductive FSL assumes all query set instances arrive simultaneously, and treat the unlabeled query set as an auxiliary set~\cite{Liu2018TPN,Li2019Learning,Chen2021ECKPN}. The relationship between unlabeled data and labeled support set facilitates the meta-model construction.

\noindent{\bf Unsupervised Meta-learning}~(UML) trains an effective meta-model without base class labels~\cite{antoniou2019assume,Metz2018Learning}.
One main obstacle is how to enable episodic training in this unsupervised scenario. By substituting ``classes'' in the base class with ``pseudo-classes'', we could easily extend supervised meta-learning to an unsupervised manner.
\cite{Hsu2018Unsupervised} utilizes off-the-shelf embedding learning methods and generates pseudo-classes by clustering multiple times, but the quality of those pseudo-labels is highly related to the results of clustering~\cite{Lee2021Meta,Xu2021Unsupervised}.
Benefiting from semantic consistency among the perturbed views of a single instance, we can treat augmented versions of an instance as if they are in the same pseudo-class.
Both embedding-based~\cite{ji2019unsupervised} and optimization-based~\cite{qin2020unsupervised} methods are investigated.
Empirical results show those direct extensions perform well and using large mini-batches yields performance improvement~\cite{medina2020self}, but a large gap still exists with their supervised counterparts~\cite{Chen2020Shot,qin2020unsupervised}.
We fill the gap between UML and its supervised upper bound by taking account of the task's characteristics in efficient episodic training. Our methods outperform current ones and get similar results with supervised methods.

\noindent{\bf Self-Supervised Learning}~(SSL) is another possible way of learning the embedding in an unsupervised manner~\cite{misra2020self}. Based on pre-text tasks without explicit usage of labels, the representation of objects becomes more discriminative and generalizable to ``downstream'' tasks~\cite{doersch2015unsupervised,noroozi2016unsupervised,pathak2016context,zhang2016colorful,gidaris2018unsupervised,ledig2017photo}. Inspired by the similarity between embedding-based meta-learning and contrastive SSL methods~\cite{he2020momentum,chen2020simple}, we can treat novel few-shot tasks as ``downstream'' ones so that the SSL-learned embeddings could help UML accordingly.
Several recent approaches apply SSL with both supervised~\cite{Su2019When} and unsupervised meta-learning~\cite{ji2019unsupervised,li2020few}, and demonstrate that an auxiliary self-supervised objective helps. 
They get a bit ``counter-intuitive'' results that SSL benefits more than episodic training, where the latter mimics the few-shot tasks in deployment during meta-training.
For example, \cite{Chen2020Shot} investigates MoCo~\cite{he2020momentum} in UML and shows MoCo outperforms many UML methods.
Although SSL methods narrow the gap between UML and the supervised methods, those specific properties of tasks captured by the episodic sampling are neglected. 
After analyzing key factors of the meta-training in UML, we enable the episodic meta-training to get discriminative embeddings by designing special sampling and similarity measure strategies. Our UML baseline and improved variants outperform the embeddings learned by SSL methods.
Experiments show that those SSL methods require deeper backbone architectures and longer training epochs to perform comparably with our methods.

\section{Preliminary of UML}
\label{sec:preliminary}
\noindent{\bf Unsupervised Meta-Learning~(UML)} for Few-Shot Learning~(FSL).
We define a {\em task} as a couple of a $K$-shot $N$-way support set $\mathcal{S}$ and a $Q$-shot $N$-way query set $\mathcal{Q}$.
The support set $\mathcal{S} = \{(\x_{i}, \y_{i})\}_{i=1}^{NK}$ contains $N$ classes and $K$ training examples per class. $\x_{i}\in\mathbb{R}^{D}$ is an instance and $\y_{i}\in \{0,1\}^N$ is its one-hot label. The query set $\mathcal{Q}$ has instances from the same distribution with $\mathcal{S}$, which is used to evaluate the classifier trained on $\mathcal{S}$.
The target of FSL is to learn a classifier from a few-shot $\mathcal{S}$ with small $K$ and to make the classifier has high discerning ability on the corresponding $\mathcal{Q}$.

In UML, we have a related {\em but unlabeled} base class set $\mathbf{B}$ (a.k.a. meta-train set). The goal of UML is to find a ``learning strategy'' such as an embedding-based classifier from $\mathbf{B}$, which could be generalized to few-shot support set with {\em non-overlapping} $N$ novel classes (a.k.a. meta-test set).
In other words, we first learn a meta-model $f$ from $\mathbf{B}$, which facilitates the construction of the task-specific classifier, \ie, $f$ predicts a query instance in $\mathcal{Q}$ conditioned on the set of $NK$ instances in $\mathcal{S}$.
The prediction rule $f(\x_j; \mathcal{S})$ is expected to generalizes to the target few-shot tasks with novel classes. 
We denote the two phases on base and novel classes --- learning and evaluating $f$ --- as meta-training and meta-test, respectively.

\noindent{\bf Episodic Sampling in UML.}
Following supervised meta-learning, UML mimics the target few-shot task via sampling episodes of pseudo tasks from $\mathbf{B}$~\cite{VinyalsBLKW16Matching,FinnAL17Model,Hsu2018Unsupervised}. Denote $\ell(\cdot, \cdot)$ as the loss function which measures the discrepancy between the prediction and the target label. We meta-learn $f$ with
\begin{equation}
	\min_f 
	\mathbb{E}_{(\mathcal{S}, \mathcal{Q})\sim\mathbf{B}}\;
	\sum_{(\x_j,\y_j)\in\mathcal{Q}}\;
	\Big[ \ell \left(f\left(\x_j;\; \mathcal{S}\right),\; \y_j\right)\Big]\;\label{eq:meta_obj}\;.
\end{equation}
By minimizing Eq.~\ref{eq:meta_obj} over sampled meta-training (pseudo) tasks, the experience in constructing an effective classifier with a few-shot support set (encoded in $f$) is expected to be transferred to meta-test tasks with novel classes.  

One key factor in UML is how to sample episodes of support and query sets from $\mathbf{B}$. In the supervised scenario, all instances in $\mathbf{B}$ has a class label, so that in each episode we randomly choose $N$ categories in $\mathbf{B}$ and $K$ (resp. $Q$) examples from each category are preserved for support set $\mathcal{S}$ (resp. query set $\mathcal{Q}$). The same sampling method cannot be applied to {\em unlabeled} $\mathbf{B}$. 
Some UML methods also utilize clustering to generate pseudo-classes for instances in $\mathbf{B}$~\cite{Hsu2018Unsupervised,Lee2021Meta}.
We use a simpler strategy that takes advantage of the semantic consistency among augmentations of images~\cite{antoniou2019assume,qin2020unsupervised,he2020momentum,chen2020simple}. 
In particular, we treat an instance and its augmentations (\eg, random crop and horizontal flip) come from the same pseudo-class, and any two instances have different pseudo-labels.
In this exemplar view, the sampled $K$-shot $N$-way support instances are the $K$ random copies of $N$ different images, and the query set contains another $Q$ copies per pseudo-class. 

Both clustering-based and augmentation-based pseudo-labeling strategies make the objective in Eq.~\ref{eq:meta_obj} biased.
For example, when we generate pseudo-labels via augmentations, although we ensure two augmented instances come from the same semantic classes, we may label semantically similar instances with different pseudo-classes~\cite{Chuang2020Debias,Huynh2020Boosting}. \cite{Saunshi2019Theoretical} proves that a rich embedding family together with some conditions may overcome the limitations of the false negative sampling. 
In practice, various self-supervised learning approaches use the augmentation-based labeling strategy to get stable and promising results~\cite{he2020momentum,chen2020simple}. Thus, we consider data augmentation for pseudo-labeling.

\noindent{\bf Similarity Measures in UML.} Another key factor in UML is the implementation of $f$ as an embedding-based classifier. 
We consider $f$ as a $d$-dimensional embedding function, \ie, $f=\phi:\mathbb{R}^D\rightarrow\mathbb{R}^d$.
Denote $y_i=n$ selecting instances in the $n$-th (pseudo) class, the corresponding class center is $\p_n=\frac{1}{K}\sum_{y_i=n} \phi(\x_i)$. The confidence of a query instance $\x_j$ belonging to the $N$ classes is based on the similarity $\mathbf{Sim}(\cdot, \cdot)$ between the query embedding and $N$ centers~\cite{SnellSZ17Prototypical}:
\begin{align}
	\hat{\y}_j = f\left(\x_j; \mathcal{S}\right) &= \mathbf{Softmax}\big(\mathbf{Sim}(\phi(\x_j), \;\p_n)\big) \label{eq:protonet} \\
	&= \left[\frac{\exp(\mathbf{Sim}(\phi(\x_j), \;\p_n))}{\sum_{n'=1}^N \exp\left(\mathbf{Sim}(\phi(\x_j), \;\p_{n'})\right)}\right]_{n=1}^N \;.\notag
\end{align}
Through the nearest class mean classifier, the larger the similarity between a query instance with a support center, the larger the probability the query instance comes from the corresponding class. $\mathbf{Sim}(\cdot, \cdot)$ in Eq.~\ref{eq:protonet} could be various metrics, for example, the negative Euclidean distance~\cite{SnellSZ17Prototypical}:
\begin{equation}
	\mathbf{Sim}_\textbf{dis}(\phi(\x_j), \;\p_n) = -\|\phi(\x_j) - \p_n\|_2^2\;,
\end{equation}
the cosine similarity~\cite{VinyalsBLKW16Matching}
\begin{equation}
	\mathbf{Sim}_\textbf{cos}(\phi(\x_j), \;\p_n) = \frac{\langle \phi(\x_j), \p_n \rangle}{\|\phi(\x_j)\|\|\p_n\|}\;,\label{eq:cosine}
\end{equation}
or the inner product
\begin{equation}
	\mathbf{Sim}_\textbf{inner}(\phi(\x_j), \;\p_n) = \langle \phi(\x_j), \p_n \rangle\;.\label{eq:inner}
\end{equation}
Different similarities have diverse effects on the meta-learned embeddings. \cite{SnellSZ17Prototypical} argues that there is a scale difference between the negative distance and cosine similarity, and \cite{Gidaris2018Dynamic,Oreshkin2018TADAM} propose to scale $\mathbf{Sim}(\cdot, \cdot)$ with an additional temperature $\tau$, \ie, $\mathbf{Sim}(\cdot, \cdot)/\tau$, to smooth the logit in Eq.~\ref{eq:protonet}. A large temperature $\tau$ pushes the query embeddings away from all non-belonging centers, while a smaller $\tau$ concentrates the force to push the closest wrongly assigned centers~\cite{Oreshkin2018TADAM,Zhang2018Heated}. 
\cite{YeHZS2018Learning} verifies the importance of $\tau$ especially when the embedding is initialized with pre-trained weights. \cite{Wang2019Simple} points out the cosine similarity with normalized embeddings works better with a pre-trained model for supervised FSL. Therefore, the way to measure instances in a task and the temperature influence the discriminative ability of the embeddings.

\begin{algorithm}[t]
	\caption{Meta-training for {\em Vanilla} UML.}
	\label{alg:vanilla_uml_meta_train}
	\begin{algorithmic}[1]{
			\REQUIRE {\textbf{Unlabeled} base class set $\mathbf{B}$}
			\FORALL{iteration = 1,...} 
			\STATE {Sample $N$ instances from $\mathbf{B}$}
			\STATE {Apply augmentations to get $N$-way $K$-shot ($\mathcal{S}$, $\mathcal{Q}$)}
			\STATE {Get $\phi(\x)$ for all $\x\in\mathcal{S}\cup\mathcal{Q}$}
			\FORALL{$(\x_j,\y_j) \in \mathcal{Q}$}
			\STATE {Get $f(\x_j, \mathcal{S})$ with selected metric}
			\STATE {Compute $\ell(f(\x_j, \mathcal{S}),\;\y_j)$}
			\ENDFOR
			\STATE {Accumulate loss for all $\x_j$ as Eq.~\ref{eq:meta_obj}}
			\STATE {Update $\phi$ with SGD}
			\ENDFOR
		}
		\RETURN {Embedding $\phi$}
	\end{algorithmic}
\end{algorithm}
\begin{algorithm}[t]
	\caption{Meta-test for UML.}
	\label{alg:vanilla_meta_test}
	\begin{algorithmic}[1]{
			\REQUIRE {A labeled support set $\mathcal{S}$ and an unlabeled query instance $\x_j$ (from {\em non-overlapping} novel class w.r.t. $\mathbf{B}$), the meta-learned embedding $\phi$}
			\STATE {Compute $\p_n$ for all $n=1,\ldots,N$ classes}
			\STATE {Compute $\mathbf{Sim}(\phi(\x_j), \;\p_n)$ with selected metric and $\phi$}
			\STATE {Predict via $\argmax_n \mathbf{Sim}(\phi(\x_j), \;\p_n)$}
			\RETURN {The predicted label of $\x_j$}
		}
	\end{algorithmic}
\end{algorithm}

\noindent{\bf Summary and discussions of UML.}
Optimizing Eq.~\ref{eq:meta_obj} with Eq.~\ref{eq:protonet} meta-learns the embedding $\phi$ in a contrastive manner --- instances are pulled to its center with the same pseudo-label, while those with different pseudo-labels are pushed away.
We find this vanilla UML method could generate semantically meaningful embeddings and helps few-shot tasks with novel ``non-pseudo'' classes, which is also validated in~\cite{khodadadeh2019unsupervised,medina2020self}.
The meta-training and meta-test phases of the vanilla UML are described in Alg.~\ref{alg:vanilla_uml_meta_train} and Alg.~\ref{alg:vanilla_meta_test}, respectively.

Inspired by the fact that self-supervised learning methods utilize data augmentation to achieve similar results with their supervised counterparts~\cite{Chen2020Big,chen2020improved}, we also expect UML to have similar properties. However, empirical results indicate there exists a large gap between UML and supervised meta-learning on FSL benchmarks~\cite{khodadadeh2019unsupervised,medina2020self}.
Despite mimicking FSL tasks, the episodic training in UML cannot introduce additional benefits when compared with ``plain'' trained embeddings with self-supervised learning ways~\cite{Chen2020Shot,li2020few}.
We find that the vanilla UML could be strong once equipped with simple modifications on key meta-training factors, facilitating filling the gap with its supervised upper bound.

\section{Analyzing Meta-Training Factors in UML}
\label{sec:analysis}
We investigate two key factors, \ie, the episodic sampling strategy and the similarity measure, in Unsupervised Meta-Learning~(UML).
With the proposed sufficient sampling and a new similarity, we get a strong UML baseline that makes UML practical for few-shot classification. 
Then we empirically explore other factors influencing UML, which provides insight into designing improved UML methods.

\subsection{Simple Modifications towards Effective UML}
We analyze how to sample episodic tasks efficiently for better gradient estimation and how to measure similarity between instances in a (pseudo) task in UML.

\noindent{\bf Sufficient Episodic Sampling~(SES).}
In Alg.~\ref{alg:vanilla_uml_meta_train}, Stochastic gradient descent (SGD) is applied to optimize the UML objective in Eq.~\ref{eq:meta_obj}, and there is one single task sampled per episode~\cite{SnellSZ17Prototypical,YeHZS2018Learning}. In other words, a gradient descent step over the embedding $\phi$ is carried out once sampling each couple of $N$-way support and query sets. 
In vanilla supervised learning, one gradient step is performed after averaging a mini-batch of losses to produce a more accurate gradient estimation. Inspired by this practical usage of SGD, we may sample multiple tasks and compute gradients over their averaged loss~\cite{Gidaris2018Dynamic,Lee2019Meta}. 
Sampling more than one task per episode accelerates the convergence of meta-training. However, embeddings of multiple tasks should be extracted accordingly via multiple forward passes, which costs high memory and the number of tasks in one episode is limited. 
\cite{FinnAL17Model,Li2019LGM} resolve the dilemma by accumulating the gradient and delaying the backward operation after multiple episodes, but cannot significantly increase the convergence speed.

We propose a more {\em efficient} implementation of the meta-training update with {\em almost no additional computational cost}. Assume we sample a $C$-way $(K+Q)$-shot task $\mathcal{S}\cup\mathcal{Q}$ in an episode where $C\ge N$, then we extract embeddings of all instances {\em with a single forward pass}. Rather than computing losses on one single pair of $\mathcal{S}$ and $\mathcal{Q}$, we randomly {\em re-split the embeddings} into couples of support and query sets. Specifically, $N$ of $C$ classes are sampled, $K$ instances from each class are randomly selected into the support set, and the remaining $Q$ instances in those $N$ classes are used for the query set. By repeating this process multiple times, we implicitly construct lots of pseudo tasks {\em in one single episode}, and one gradient descent is executed for averaged ``multi-task'' losses.
The main steps of SES are in Alg.~\ref{alg:baseline_meta_train} lines 5-6 and Fig.~\ref{fig:method} (a). Since only one forward of $\phi$ is applied (Alg.~\ref{alg:baseline_meta_train}, line 4), SES has negligible additional computational costs during meta-training. 
In our implementation, we set $C=N$, so our baseline has the same mini-batch size as the vanilla method.
Experiments show SES improves the efficiency of meta-training, which fully utilizes instances in a mini-batch.

\noindent{\bf Semi-Normalized Similarity~(SNS).}
The similarity metric in Eq.~\ref{eq:protonet} plays an essential role, which determines whether two instances are similar in a task. As mentioned in~\cite{Gidaris2018Dynamic,Oreshkin2018TADAM,YeHZS2018Learning} and Section~\ref{sec:preliminary}, the temperature $\tau$ in $\mathbf{Sim}(\cdot, \cdot)/\tau$ smooths the logit and influences the discriminative ability of the meta-learned embedding. 
Instead of manually setting $\tau$, we propose to {\em learn} the temperature adaptively during the meta-training progress. Moreover, we make the $\tau$ instance-specific, denoted as $\tau_\x$, when measuring the similarity between $\x$ and prototypes. 
We take advantage of the fact that {\em scale} of the embedding, \ie, the embedding norm $\|\phi(\x)\|$, will not influence the relative similarity between instances in Eq.~\ref{eq:protonet}, so $\phi$ will have a freedom degree on its scale. 
With the help of the embedding norm, we propose to obtain $\tau_\x$ as a function of $\|\phi(\x)\|$, which also avoids introducing additional learnable parameters to determine $\tau_\x$. 
The model will adapt $\|\phi(\x)\|$ as well as $\tau_\x$ to the right scale and dynamically improve the discriminative ability of the learned embeddings.
 
To simplify the form, we implement the temperature as the reciprocal of the embedding norm, \ie, $\tau_\x = 1/\|\phi(\x)\|$.\footnote{We omit the case $\|\phi(\x)\|=0$ since $\phi(\x)$ gets all zero elements with almost no chance.} Thus, together with the cosine similarity, we get our SNS as 
\begin{align}
	&\mathbf{Sim}(\phi(\x_j), \;\p_n) \;=\; \mathbf{Sim}_{\textbf{cos}}(\phi(\x_j), \;\p_n) / \tau_{\x_j} \;\notag\\
	\;=\;\;\;&\mathbf{Sim}_{\textbf{cos}}(\phi(\x_j), \;\p_n) \cdot \|\phi(\x_j)\| = \phi(\x_j)^\top \frac{\p_n}{\|\p_n\|_2}\;\label{eq:protomatch}.
\end{align}
The name ``semi-normalized'' comes from the fact that SNS only normalizes support centers in Eq.~\ref{eq:protomatch}.
Recall that the similarity between a query instance $\x_j$ and multiple support centers $\{\p_n\}$ are compared {\em simultaneously} with the softmax operator, missing the normalization on $\phi(\x_j)$ in SNS leads to the {\em same} prediction with the cosine similarity in Eq.~\ref{eq:cosine}. 

\begin{figure}[t]
	\centering
	\includegraphics[width=0.9\linewidth]{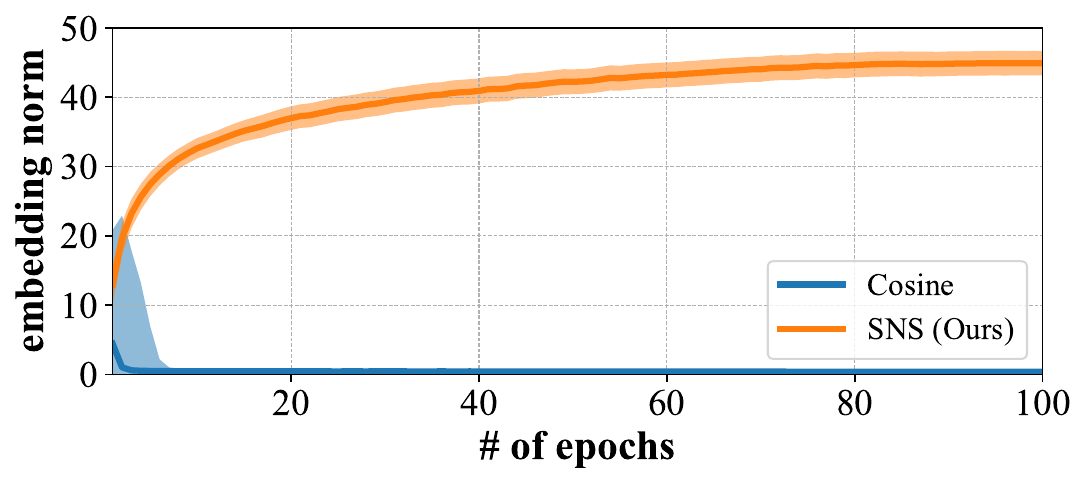}
	\caption{The change of the embedding norm $\|\phi(\x)\|$ (mean as well as std. over all instances) during the meta-training progress when learning with cosine similarity and SNS on {\it Mini}ImageNet. We set $\phi$ as a four-layer ConvNet. The detailed setup is the same as that described in Section~\ref{sec:analysis_experiments}.}\label{fig:emb_norm}
\end{figure}

Since the embedding norm of an instance decreases when we use $\mathbf{Sim}$ as the cosine similarity during the meta-training progress (as shown in Fig.~\ref{fig:emb_norm}), the instance-specific temperature in SNS becomes larger gradually, which softens the logits in an {\em adaptive} manner.
In detail, based on the discussions in Section~\ref{sec:preliminary}, the loss forces to push a query instance far from nearest non-belonging centers at the initial optimization stage (with relative larger $\|\phi(\x_j)\|$), which captures the {\em local} similarity relationship in a task. When the embedding norm becomes smaller, the temperature helps concentrate the gradient from pushing nearest impostor centers to all non-belonging ones, which makes the optimization focuses on the {\em global} property among tasks.
Experiments verify that SNS facilitates learning the best temperature for $\mathbf{Sim}_\textbf{cos}$ and improves UML on various configurations of tasks. 

\newtheorem{remark}{Remark}

We summarize our UML baseline with the help of SES and SNS, whose meta-training workflow is listed in Alg.~\ref{alg:baseline_meta_train} (especially lines 5, 6, and 8). {\em No additional parameters are introduced in our UML baseline}.

\begin{algorithm}[t]
	\caption{The meta-training flow of our UML Baseline.}
	\label{alg:baseline_meta_train}
	\begin{algorithmic}[1]{
			\REQUIRE {Unlabeled base class set $\mathbf{B}$}
			\FORALL{iteration = 1,...} 
			\STATE {Sample $C$ instances from $\mathbf{B}$}
			\STATE {Apply $K+Q$ augmentations for all $C$ instances}
			\STATE {Get $\phi(\x)$ for all augmentations}
			\FORALL{{\color{blue}{task\_iter = 1,...}}}
			{\color{blue}\STATE {Split embeddings to get a task ($\mathcal{S}$, $\mathcal{Q}$)}}
			\FORALL{$(\x_j,\y_j) \in \mathcal{Q}$}
			{\color{blue}\STATE {Get $f(\x_j, \mathcal{S})$ {with SNS} in Eq.~\ref{eq:protomatch}}}
			\STATE {Compute $\ell(f(\x_j, \mathcal{S}), \;\y_j)$}
			\ENDFOR
			\ENDFOR
			\STATE {Accumulate loss as Eq.~\ref{eq:meta_obj} for all tasks}
			\STATE {Update $\phi$ with SGD}
			\ENDFOR
		}
		\RETURN {Embedding $\phi$}
	\end{algorithmic}
\end{algorithm}

\subsection{Empirical Analyses of Key Factors for UML}\label{sec:analysis_experiments}
Before we move on, we analyze the effectiveness of our UML baseline and explore other key elements in UML's meta-training, which lays the foundations for designing improved UML methods.
Following the configuration in~\cite{khodadadeh2019unsupervised}, we implement $\phi$ with a four-layer ConvNet and focus on the {\it Mini}ImageNet~\cite{VinyalsBLKW16Matching} benchmark with the standard split~\cite{Sachin2017}. During meta-training, the initial learning rate is set to 0.002 and is cosine annealed over 100 epochs. $C=64$ instances are sampled in each mini-batch as different pseudo-classes, and each instance is augmented into six copies.
We set $N=64$, $K=1$, $Q=5$ by default. 
During meta-test, we evaluate the meta-learned embeddings over 10,000 few-shot tasks (where $N=5$ and $Q=15$) from novel classes.
Note that we only use labels to evaluate statistics such as few-shot classification accuracy, and {\em no base class labels are utilized during meta-training}.
We observe the same phenomenon on other datasets with deeper backbones. More results and analyses are in Section~\ref{sec:experiment}.
Other configurations of episodes and their influences on UML are in the supplementary.

\begin{figure*}
	\begin{minipage}{0.45\linewidth}
		\centering
		\includegraphics[width=\textwidth]{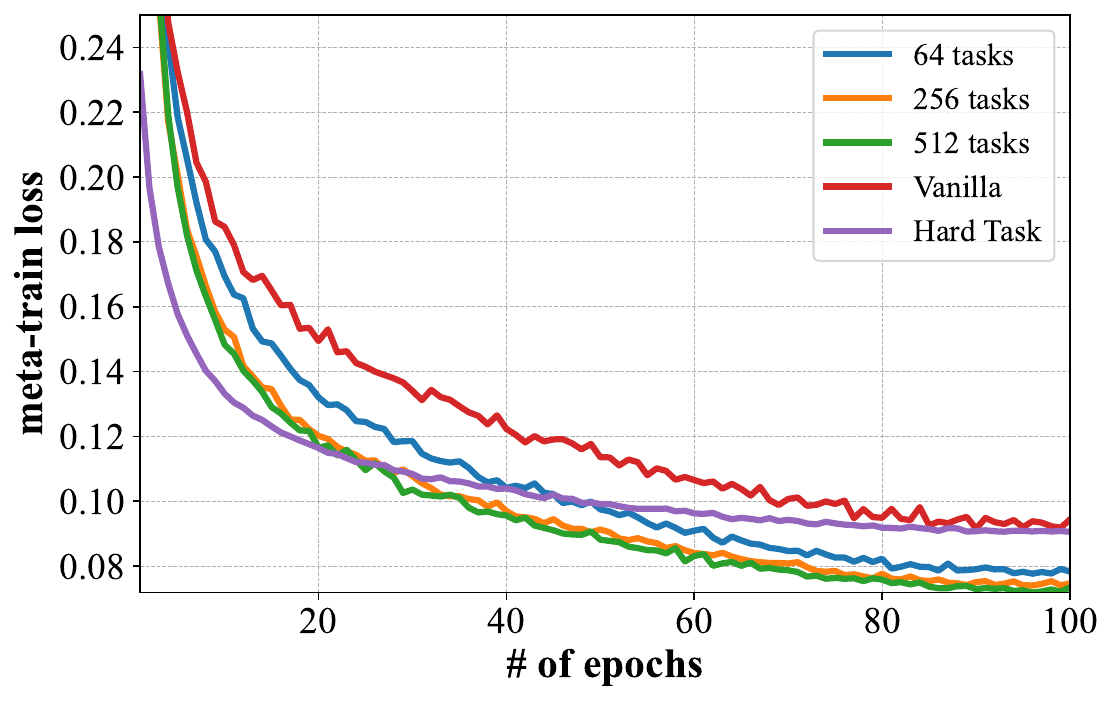}
	\end{minipage}
	\begin{minipage}{0.45\linewidth}
		\centering
		\includegraphics[width=\textwidth]{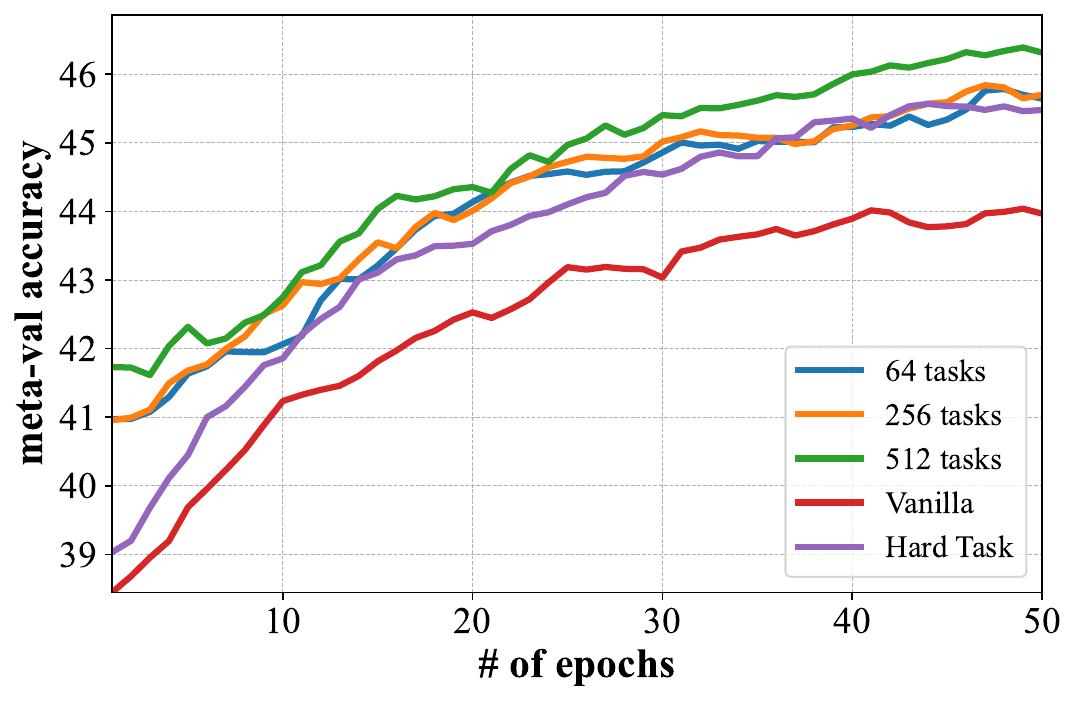}
	\end{minipage}
	
	\begin{minipage}{0.43\linewidth}
		\centering
		\includegraphics[width=\textwidth]{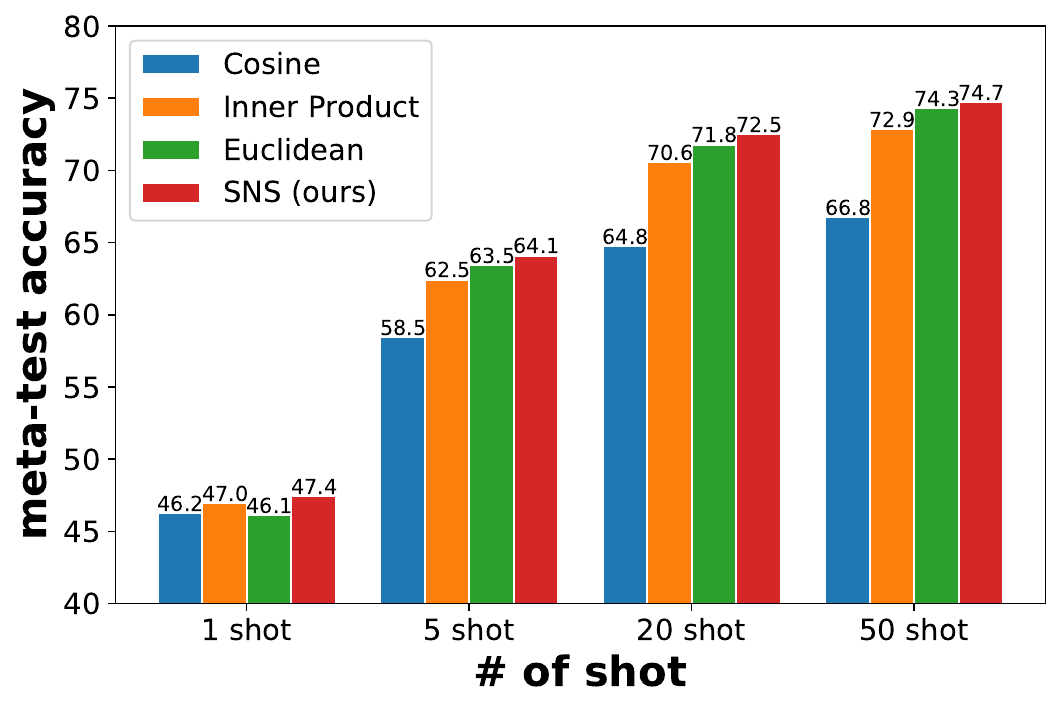}
	\end{minipage}
	\begin{minipage}{0.45\linewidth}
		\centering
		\includegraphics[width=\textwidth]{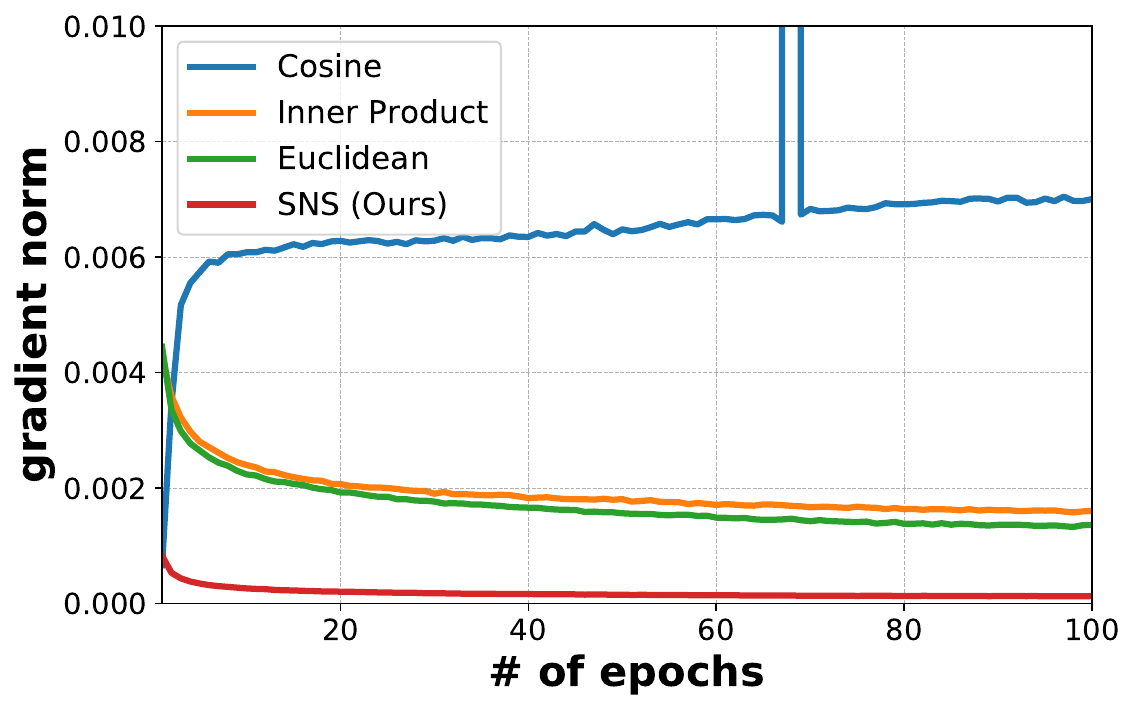}
	\end{minipage}
	\vspace{-10pt}
	\caption{Empirical study to show the importance of sufficient sampling and semi-normalized similarity in UML on {\it Mini}ImageNet with a four-layer ConvNet over 100 epochs. {\it Upper}: the meta-training loss and 5-way 1-shot meta-validation accuracy along epochs. Sufficient sampling with multiple tasks demonstrates fast convergence speed and high generalization ability. 
	{\it Lower left}: By comparing different similarity measures, we find SNS always performs the best when tested with 5-way $\{1,5,20,50\}$-shot tasks.
	{\it Lower right}: The gradient norms w.r.t. embeddings averaged over all instances in the mini-batch along with the meta-training progress. Different similarities have diverse changes in their gradient norms.
	}
	\label{fig:num_task}
\end{figure*}

\noindent{\bf Does sufficient sampling with SES help UML?}
We evaluate SES by gradually increasing the number of re-sampled tasks in one episode from 1 (denoted as the ``vanilla'' case) to 512, and we implement the similarity measure in Eq.~\ref{eq:protonet} with our SNS.
The change of meta-training loss and meta-validation 1-shot 5-way accuracy over sampled 10,000 tasks are plotted in Fig.~\ref{fig:num_task} (upper). 
During the meta-training progress, the loss of UML objective in Eq.~\ref{eq:meta_obj} decreases consistently, which indicates that constructing pseudo-classes with augmented views makes the meta-learned embedding discriminative. 
Obviously, the model converges faster and generalizes better by re-sampling more tasks in one episode. There is a huge gap over the meta-training loss/meta-validation accuracy between the 512-task case and the vanilla one, which verifies the importance of sampling sufficient tasks in meta-training. 
We find that monotonously increasing the task number has no additional improvements. So in the following experiments, we set the number of tasks in SES to 512. 

We also compare SES with  ``Hard Task'' sampling~\cite{Sun2019Meta,Sun2020Meta} (denoted as ``HT'' for short). HT selects hard classes from various tasks and then organizes them together. We find HT accelerates meta-training at first but slows down later. 
The reason mainly comes from HT's dependence on semantic information to construct hard tasks, which becomes difficult in UML. 
SES achieves a faster convergence rate and higher meta-val accuracy than HT once with enough tasks.

\noindent{\bf Does the new similarity SNS help UML?}
We compare cosine similarity, (negative) euclidean distance, inner product, and SNS. We meta-train those similarities with SES and evaluate the learned embedding with the corresponding similarity over 5-way $\{1,5,20,50\}$-shot novel class tasks. 
We set the default temperature for cosine as 0.5 and 1 for others. 
The FSL accuracy on the meta-test set is shown in Fig.~\ref{fig:num_task} (lower left). Different from supervised meta-training where cosine is the best choice~\cite{Wang2019Simple}, in UML, later similarities outperform the former one in more shot scenarios. 
SNS performs the best when the embedding is meta-trained with one task per episode (results are in the supplementary). When equipped with SES, SNS also shows {\em stable} improvements when evaluated with different configurations of tasks.

Besides the change of embedding norm in Fig.~\ref{fig:emb_norm}, we analyze the norm of the embedding gradient in Fig.~\ref{fig:num_task} (lower right), \ie, $\|\nabla_{\phi(\x)} \ell\|_2$ averaged over all instances in the mini-batch, along with the meta-training progress. 
We observe that cosine has very small embedding norms, so normalizing with which results in large gradient norms. The large gradients could be one reason for its instability in UML, which requires carefully tuned temperatures. 
Inner product and Euclidean distance without embedding normalization lead to relatively smaller gradient norms.
Our SNS, however, has the smallest gradient norm compared with others. We conjecture that $\phi$ is updated in diverse ways by using different similarity measures. SNS makes $\phi$ approach the stable point directly, while others such as inner product make aggressive updates at first and then gradually adapt the embedding to the target solution in a zigzag way. Based on the meta-test classification performance, SNS converges faster and generalizes better.

\begin{figure}[t]
    \centering
    \includegraphics[width=0.9\linewidth]{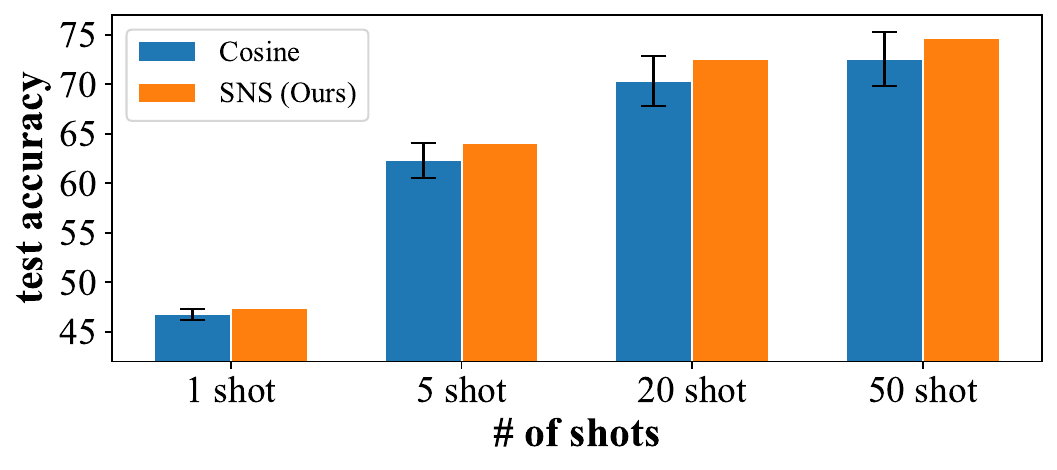}
    \caption{Comparison between SNS and cosine on {\it Mini}ImageNet with ConvNet. We carefully tune the temperature for cosine with 16 values in $[0.005, 1]$, and show mean as well as std. over all temperatures. }
    \label{fig:temperature_var}
\end{figure}

We also manually tune the temperature for cosine over 16 values $\tau\in[0.005, 1]$.
We observe that the learned embeddings show diverse performance when trained with various temperatures, and the best $\tau$ for different task configurations (\eg, different $K$) varies. 
The mean and std. over temperatures in Fig.~\ref{fig:temperature_var} indicate large variances, and SNS performs on par with the best-performed cosine similarity in all cases {\em with default $\tau=1$}.
Since we cannot determine the best $\tau$ for cosine with the final performance in advance, SNS is more {\em practical}.

\begin{figure}[t]
	\centering
	\includegraphics[width=0.9\linewidth]{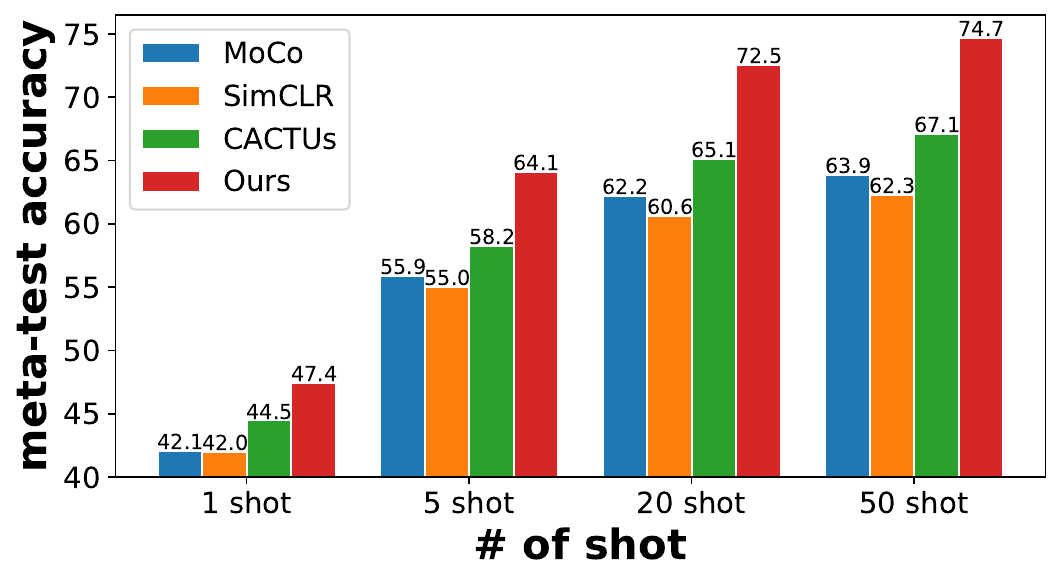}
	\caption{Comparisons among our augmentation-based UML baseline, clustering-based UML method CACTUs, and representative self-supervised learning methods (\ie, MoCo and SimCLR). All methods are trained with a four-layer ConvNet on {\it Mini}ImageNet. The meta-learned embeddings with UML show better discriminative ability on ``downstream'' novel class 5-way $\{1,5,20,50\}$-shot tasks.}\label{fig:SSL_comparison}
\end{figure}

\noindent{\bf How to generate pseudo-labels --- clustering or augmentations?} We investigate the difference between two pseudo-labeling strategies. We mainly consider the clustering choice CACTUs in~\cite{Hsu2018Unsupervised}. We find that the labeling quality of clustering highly depends on the off-the-shelf learned embedding. We try several embedding learning methods (including those in \cite{Hsu2018Unsupervised} and self-supervised learning ones) for clustering and show the best-performed CACTUs in Fig.~\ref{fig:SSL_comparison}. We also equip CACTUs with SES for fair comparisons. We find our UML baseline consistently outperforms CACTUs, which indicates data augmentation could be a stable and efficient manner to generate pseudo-labels in UML.

\begin{table}[tbp]
	\centering
	\caption{Comparison of different augmentations on {\it Mini}ImageNet with ConvNet backbone. Methods are evaluated over different $N$-way $K$-shot tasks. Compared augmentations are Simple, SimCLR, AMDIM, AutoAug, and RandAug. ``Simple'' consists of random resized crop, color distortion, and random horizontal flip, which is commonly used in supervised learning. Moderate augmentation such as AMDIM shows the best results. Too strong augmentation like RandAug will degenerate the performance.}
	\begin{tabular}{lcccc}
		\addlinespace
		\toprule
{\bf ($N$, $K$)} & \textbf{(5,1) }                & \textbf{(5,5) }                & \textbf{(5,20) }               & \textbf{(5,50) }               \\
\midrule
Simple          & 44.42         & 60.18      & 69.07        & 71.63         \\
SimCLR~\cite{chen2020simple}          & 46.51          & 62.91          & 71.35         & 73.66          \\
AMDIM~\cite{Bachman2019Learning}           & \textbf{47.43} & \textbf{64.11} & \textbf{72.52} & \textbf{74.72} \\
AutoAug~\cite{Cubuk2018Auto}         & 44.81          & 61.05         & 69.94          & 72.62         \\
RandAug~\cite{Cubuk2020Rand}         & 45.94          & 61.47          & 68.85         & 71.12         \\
		\bottomrule
	\end{tabular}
	\label{tab:augment_conv}
		\vspace{-4pt}
\end{table}

\noindent{\bf Moderate augmentation leads to better performance.}
Data augmentation plays an important role in contrastive self-supervised learning~\cite{he2020momentum,chen2020simple,tian2019contrastive}. We investigate how augmentation influences the performance of our UML baseline. We compare augmentations from SimCLR~\cite{chen2020simple}, AMDIM~\cite{Bachman2019Learning}, AutoAug~\cite{Cubuk2018Auto} and RandAug~\cite{Cubuk2020Rand}. We also compare an augmentation ``Simple'' that consists of random resized crop, color distortion, and random horizontal flip. Results show that relatively strong augmentation will benefit our UML baseline. However, too strong augmentation such as RandAug may degrade the performance. These results are in accord with those in self-supervised learning~\cite{Tian2020What}. In our experiments, AMDIM is the best augmentation.

\noindent{\bf Episodic training works better than contrastive Self-Supervised Learning (SSL) baselines.}
An intuitive question is whether episodic training outperforms the contrastive SSL methods. 
We compare our UML baseline with two representative SSL methods, \ie, MoCo~\cite{he2020momentum} and SimCLR~\cite{chen2020simple}. Hyper-parameters are tuned carefully for SSL methods on the meta-validation set, and they require longer epochs to converge (\eg, 800 epochs).
Fig.~\ref{fig:SSL_comparison} shows the superiority of our episodic meta-trained UML baseline when compared with the ``plain'' trained SSL methods. The results indicate episodic training is an essential factor for UML. 

\noindent{\bf Summary.}
Benefiting from the task sampling strategy (\ie, SES) and the in-task similarity measure (\ie, SNS), our UML baseline meta-learns generalizable embeddings without base class labels. 
The results indicate the specific consideration of task characteristics with episodic training is essential for UML, so we keep SES and SNS as default configurations for UML and propose to design improved UML approaches.

\section{Further Exploring Characteristics of Tasks for UML}
\label{sec:method}
Analyses in Section~\ref{sec:analysis} verify the effectiveness of several key factors in UML, including generating pseudo-labels with data augmentations, episodic sampling tasks with SES, and measuring in-task similarities with SNS.
We further explore the characteristics of tasks to improve the discriminative and generalization ability of UML. Due to the diversity of sampled tasks in meta-training, we propose either ``amplifying'' or ``resolving'' the task level differences, which achieve stronger {\em lower-shot} (\eg, 1-shot) and {\em higher-shot} (\eg, 20-shot) FSL performance, respectively. 

\begin{figure*}
	\centering
	\includegraphics[width=1\linewidth]{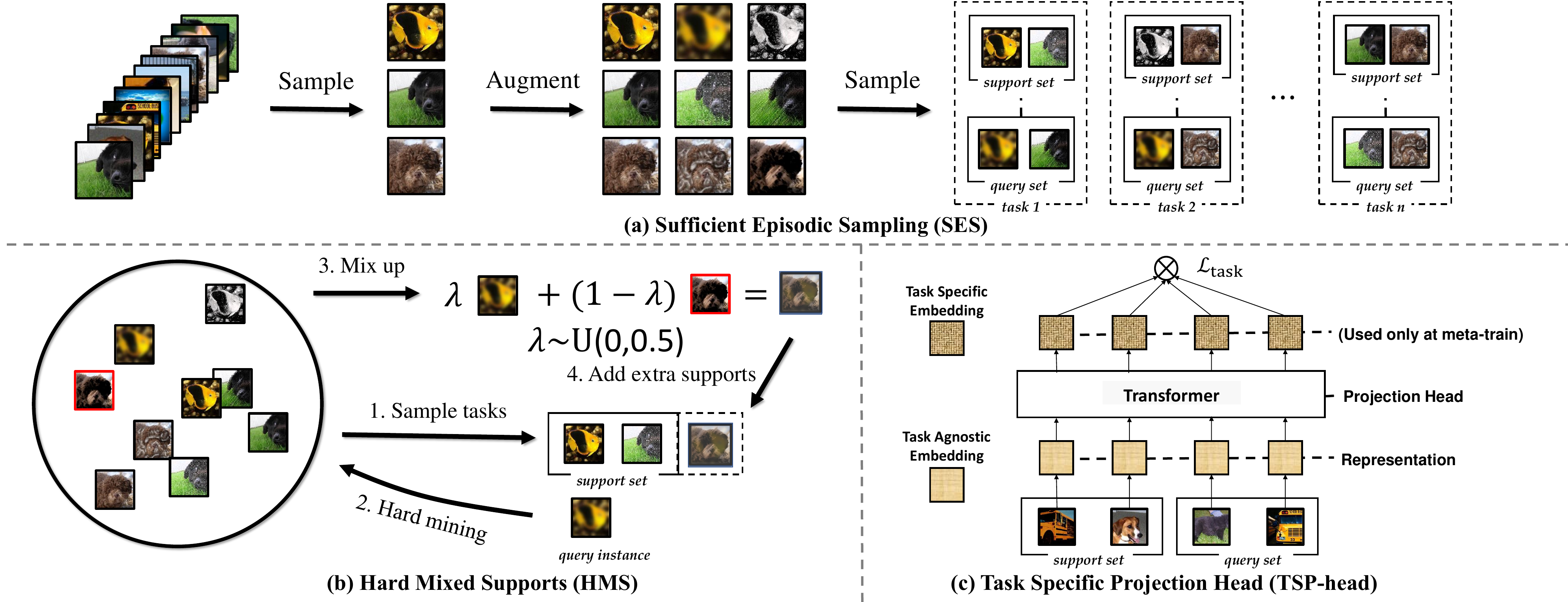}
	\caption{\textbf{Illustrations of our proposed UML methods.} ({\it a}) Based on a sampled mini-batch, we apply image augmentations to construct pseudo-classes. Multiple tasks are re-sampled from a mini-batch with Sufficient Episodic Sampling~(SES). 
	({\it b}) Given a query embedding, we mix it up with the most similar instances (with mixup strength coefficient $s$). The mixed embeddings are then added as extra supports for this query, which constructs a query-specific difficult task. 
	({\it c}) To relieve the variety of task distributions, we transform task agnostic embeddings to task-specific ones with an auxiliary task-specific projection head (implemented with Transformer~\cite{Vaswani2017Attention}) during meta-training. This component is {\em discarded} during meta-test.}\label{fig:method}
\end{figure*}

\subsection{Adaptive Difficult Tasks via Hard Mixed Supports}
In our vanilla UML baseline, all (pseudo) tasks during meta-training are sampled uniformly.
We try to amplify the characteristics of tasks by learning more confusing tasks, \eg, tasks discerning ``hound'' and ``husky'' rather than ``dog'' and ``cat''. Those confusing tasks facilitate the embedding-based classifier to identify key features to differentiate the neighbors and impostors. 
Then, the embedding could be more discriminative and helps novel class FSL tasks. 
However, given the unlabeled base class set, we cannot collect semantically related classes in one task directly, so we choose to measure the difficulty of a task via the similarity in the embedding space --- between a query instance and the support centers from those nearest non-belonging classes.

We propose Hard Mixed Supports (HMS) to meta-learn more discriminative embeddings. 
HMS dynamically constructs more difficult tasks with support distractors. 
Given a couple of support and query sets $(\mathcal{S}, \mathcal{Q})$, we synthesize hard distractors for each query instance $\x_j\in\mathcal{Q}$ by mixing-up $\x_j$ with its nearest neighbor $\x_i\in\mathcal{S} \cup \mathcal{Q}$ coming from different pseudo-classes. Formally, we find $K$ nearest neighbors of $\x_j$ in the embedding space:
\begin{equation}
	\hat{\mathcal{S}}_j = \underset{\x_i\in\mathcal{S} \cup \mathcal{Q}}{\mathrm{argmax}_K}\left\{\mathbf{Sim}\left(\phi(\x_j),\;\phi(\x_i)\right),\;y_j \neq y_i \;\right\}\;.
\end{equation}
$\mathbf{Sim}(\cdot,\cdot)$ is SNS as Eq.~\ref{eq:protomatch}. $y_i$ and $y_j$ are the pseudo-classes of $\x_i$ and $\x_j$, respectively. $\mathrm{argmax}_K$ selects the top-$K$ instances with the highest similarity~\cite{Schroff2015FaceNet}, \ie, the most similar ones. To increase the difficulty, we further mixup $\phi(\x_j)$ with the embeddings of $\hat{\mathcal{S}}_j$:
\begin{equation}
	\tilde{\mathcal{S}}_j = \left\{ \lambda \phi(\x_j) + (1-\lambda) \phi(\x) \;|\;\x\in \hat{\mathcal{S}}_j\right\}\label{eq:mixup}\;.
\end{equation}
$\lambda\in[0, 0.5]$ is a random value sampled from a uniform distribution.
Eq.~\ref{eq:mixup} interpolates distractors between the query instance embedding $\phi(\x_j)$ with the embedding of its hard mined neighbor $\phi(\x)$. 
The strength of the mixup coefficient $\lambda$ is controlled so that the mixed instances are biased towards the mined neighbor, which guarantees the semantic space is not messed up. But in experiments, we find a larger range of $\lambda$ (\eg, $\lambda\in[0, 1]$) facilitates 1-shot meta-test tasks.
Each instance in the mixed support set $\tilde{\mathcal{S}}_j$ is taken as a new pseudo-class other than those in $\mathcal{S}$. The confusing $\tilde{\mathcal{S}}_j$ amplifies the difficulty of discriminating the right queries.
Finally, we augment the original support set $\mathcal{S}$ with the mixed one $\tilde{\mathcal{S}}_j$ and obtain a higher-way confusing support set. Denote the specific support embedding set $\mathcal{S}_j = \{\phi(\x_i)\;|\;\x_i\in\mathcal{S}\} \cup \tilde{\mathcal{S}}_j$ for each query instance $\x_j$, we re-compute the loss function in Eq.~\ref{eq:meta_obj} over $\mathcal{S}_j$ and optimize the model with back-propagation. HMS is operated over the embeddings, so it
incurs a negligible computational burden.

In summary, we construct a query-specific hard support set by augmenting the support set with distractors for each query instance. The more confusing support set amplifies the characteristic of tasks, which leads to more discriminative embeddings.
The main flow of HMS is listed in Alg.~\ref{alg:hms} (the changes w.r.t. our baseline are in lines 8-13). 
\begin{algorithm}[t]
	\caption{The meta-training flow of HMS.}
	\label{alg:hms}
	\begin{algorithmic}[1]{
			\REQUIRE {Unlabeled base class set $\mathbf{B}$}
			\FORALL{iteration = 1,...} 
			\STATE {Sample $C$ instances from $\mathbf{B}$}
			\STATE {Apply $K+Q$ augmentations for all $C$ instances}
			\STATE {Get $\phi(\x)$ for all augmentations}
			\FORALL{{task\_iter = 1,...}}
			\STATE {{Split instances to get an $N$-way $K$-shot task ($\mathcal{S}$, $\mathcal{Q}$)}}
			\FORALL{$(\x_j,\y_j) \in \mathcal{Q}$}
			{\color{blue}\STATE {{Get $\hat{\mathcal{S}}_j = \underset{\x_i\in\mathcal{S} \cup \mathcal{Q}}{\mathrm{argmax}_K}\left\{\mathbf{Sim}\left(\phi(\x_j),\;\phi(\x_i)\right) ,y_j \neq y_i \;\right\}$}}
			\STATE {{Sample $\lambda \sim U(0,0.5)$}}
			\STATE {{$\tilde{\mathcal{S}}_j = \left\{\lambda \phi(\x_j) + (1-\lambda) \phi(\x)\;|\;\x\in \hat{\mathcal{S}}_j\right\}$}}
			\STATE {{Get $\mathcal{S}_j = \{\phi(\x_i)\;|\;\x_i\in\mathcal{S}\} \cup \tilde{\mathcal{S}}_j$}}
			\STATE {Get $f(\x_j, {\mathcal{S}_j})$ {with SNS} in Eq.~\ref{eq:protomatch}}
			\STATE {Compute $\ell(f(\x_j, {\mathcal{S}_j}), \y_j)$}}
			\ENDFOR
			\ENDFOR
			\STATE {Accumulate loss as Eq.~\ref{eq:meta_obj} for all tasks}
			\STATE {Update $\phi$ with SGD}
			\ENDFOR
		}
		\RETURN {Embedding $\phi$}
	\end{algorithmic}
\end{algorithm}

\noindent{\bf Discussions.}
Selecting Hard Tasks (HT) has been verified to be effective in supervised meta-learning~\cite{Sun2019Meta,Sun2020Meta} and makes the meta-model robust. 
The hard classes selected from various tasks in HT may not increase the discerning difficulty of their combinations. 
HMS directly synthesizes confusing embeddings towards query-specific difficult tasks without multiple forward passes as in~\cite{Sun2019Meta,Sun2020Meta}, which is efficient and effective for UML.
To demonstrate the benefit of HMS, we compare HMS with HT and various meta-learning mixup variants~\cite{Xu2021Unsupervised,Yao21Improving} in Section~\ref{sec:experiment}.

Similar ideas constructing hard negatives have also been explored in self-supervised learning~(SSL)~\cite{Kalantidis2020Hard,Shen2020UnMix,Varamesh2020Mix,Ye2020Augmentation}, where the main obstacle is to ensure the synthesized negatives are confusing. For example, MoCHi~\cite{Kalantidis2020Hard} uses a large memory bank to guarantee diverse negative candidates could be searched. 
Different from adding negative examples in the contrastive loss, HMS augments the support set with confusing embeddings labeled as new pseudo-classes. Besides, HMS utilizes SES and searches for distractors in a mini-batch efficiently without losing the diversity of candidates. 
Experiments show HMS outperforms SSL methods such as MoCHi when evaluated on 1-shot tasks. We also investigate the influence of SES on HMS in the supplementary.

\subsection{Task-specific Projection Head}\label{sec:tsp_head}
The embedding $\phi$ could have diverse (or even contradictory) update directions during meta-training since the optimal embedding-based classifiers for sampled tasks are different. 
Moreover, the tasks during meta-training and meta-test of UML are composed of pseudo-classes and real semantic classes, respectively. So the distribution gap makes the embedding learned on the unlabeled base classes hard to fit the tasks from novel classes.

Unlike HMS, we turn to resolve the negative effect owing to the characteristics of pseudo UML tasks.
We propose to adapt embeddings in a task with an auxiliary Task-Specific Projection Head (TSP-Head)~\cite{YeHZS2018Learning}, which decouples the specific and generalizable components --- the specific properties in a (pseudo) task will be captured by the top-layer transformation, while the pre-adapted embedding $\phi$ could be more generalizable and facilitates novel class few-shot tasks during meta-test.

In meta-training, support and query embeddings in a task are transformed with an auxiliary set-to-set function $\mathbf{T}$:
\begin{equation}
	\psi(\x) = \mathbf{T}(\phi(\x) \;|\; \x \in \mathcal{S}\cup\mathcal{Q})\;.\label{eq:tsp_adapt}
\end{equation}
$\mathbf{T}(\cdot)$ contextualizes the union of support and query sets.
We measure the similarity between query and support in Eq.~\ref{eq:protomatch} with the transformed task-specific embedding $\psi(\x)$ {\em in meta-training}, while during meta-test only $\phi(\x)$ is used.

Following~\cite{YeHZS2018Learning}, we implement $\mathbf{T}(\cdot)$ with Transformer~\cite{Vaswani2017Attention}, and $\phi(\x)$ is adapted based on a ``key-value'' dictionary module. In particular, denote $W_Q$, $W_K$, and $W_V$ as three $d\times d$ projections, and the affinity $\alpha_i$ between one instance $\x$ and another instance $\x_i$ in a sampled task is measured by their projected inner product
\begin{equation}
	\alpha_i = \mathbf{Softmax}_i\left(\frac{\left(W_Q^\top \phi(\x)\right)^\top \left(W_K^\top \phi(\x_i)\right)}{\sqrt{d}}\right)\;.
\end{equation}
The transformed embedding $\psi(\x)$ is a weighted sum over another transformed set of embeddings $\{\phi(\x)\}_{\x \in \mathcal{S}\cup\mathcal{Q}}$ in the union of support and query sets
\begin{equation}
	\psi(\x) = \phi(\x) + \mathbf{L}\left(\sum_{\x\in \mathcal{S}\cup\mathcal{Q}} \alpha_i \left(W_Q^\top \phi(\x)\right)\right)\;.
	\label{eq:trans}
\end{equation}
$\mathbf{L}(\cdot)$ is a sequential operation of layer normalization~\cite{Ba2016Layer}, dropout~\cite{Srivastava2014Dropout}, and linear projection. 
There are two extension configurations for Transformer. First, $\mathbf{T}$ could be processed multiple times (a.k.a. multi-layer). Second, if more than one set of projection matrices are allocated, multiple adapted embeddings could be concatenated followed by a linear projection to dimensionality $d$ (a.k.a. multi-head). In our empirical study, we find the multi-head version of Transformer works better, but using more layers will not help. Detailed results are in the supplementary.

Alg.~\ref{alg:TSP} shows the main flow of TSP-Head (the changes w.r.t. our UML baseline model are in lines 7, 9, and 14).
Based on Eq.~\ref{eq:tsp_adapt}, the adapted embedding $\psi$ takes a holistic consideration of other embeddings in the task, which frees the vanilla embedding $\phi$ from encoding the specific properties of different tasks. In other words, with the help of $\psi$ dealing with characteristics of tasks, $\phi$ becomes more generalizable, so we apply $\phi$ to measure the similarity in evaluation. 

TSP-Head has another two benefits. 
It is inevitable that we assign semantically similar instances into different pseudo-classes in UML tasks, and requiring the embedding to distinguish between them will mislead models to push instances from the same semantic category away. TSP-Head accounts for possible semantic contradiction with the adapted embeddings and counters the negative influence of the semantically similar instances from different pseudo-classes.
Furthermore, UML suffers from the shift of task distribution between base and novel classes. The tasks in meta-training are sampled from an unlabeled base class set with augmentations, while a novel few-shot support set during meta-test is composed according to supervised semantic labels.
This difference indicates that we could not directly adapt task-specific embedding as in the supervised FSL, and a more generalizable $\phi$ with the auxiliary $\mathbf{T}$ helps.

\begin{algorithm}[t]
	\caption{The meta-training flow of TSP-Head.}
	\label{alg:TSP}
	\begin{algorithmic}[1]{
			\REQUIRE {Unlabeled base class set $\mathbf{B}$}
			\FORALL{iteration = 1,...} 
			\STATE {Sample $C$ instances from $\mathbf{B}$}
			\STATE {Apply $K+Q$ augmentations for all $C$ instances}
			\STATE {Get $\phi(\x)$ for all augmentations}
			\FORALL{{task\_iter = 1,...}}
			\STATE {{Split instances to get an $N$-way $K$-shot task ($\mathcal{S}$, $\mathcal{Q}$)}}
			{\color{blue}\STATE {{Transform embedding in the task\\ $\psi(\x) = \mathbf{T}(\phi(\x) \;|\; \x \in \mathcal{S}\cup\mathcal{Q})$}}}
			\FORALL{$(\x_j,\y_j) \in \mathcal{Q}$}
			{\color{blue}\STATE {Get $f(\x_j, \mathcal{S})$ {with SNS} {  and $\psi(\x)$} in Eq.~\ref{eq:protomatch}}}
			\STATE {Compute $\ell(f(\x_j, \mathcal{S}), \y_j)$}
			\ENDFOR
			\ENDFOR
			\STATE {Accumulate loss as Eq.~\ref{eq:meta_obj} for all tasks}
			{\color{blue}\STATE {Update $\phi$ and $\mathbf{T}$ with SGD}}
			\ENDFOR
		}
		\RETURN {Embedding $\phi$}
	\end{algorithmic}
\end{algorithm}

\noindent{\bf Discussions.}	
Transformer is used as an adaptation component in the meta-model to obtain task-specific embeddings in FEAT~\cite{YeHZS2018Learning}, where the adapted embeddings are applied for both base and novel tasks.
However, in TSP-Head we use Transformer as an auxiliary component to take account of characteristics of different (pseudo) tasks, \ie, we compute loss with adapted embedding during meta-training and only the vanilla embedding $\phi$ is used during meta-test.
Experiments show that using the adapted embedding $\psi$ during meta-test suffers from the distribution gap since those tasks are not generated based on data augmentations. 
We also find that Transformer performs differently in supervised and unsupervised cases. For example, a multi-head version of $\mathbf{T}$ cannot help FEAT but is useful in our TSP-Head.
	
The projection head is a commonly used trick in self-supervised learning~\cite{chen2020simple,Chen2020Big} to improve the quality of the representation. In detail, another linear layer is jointly learned in training and is discarded during evaluation. 
Different from the vanilla supervised training where all instances are {\em i.i.d.} sampled, in UML, instances in a task are related. Thus the set-wise diversity in UML is difficult to be captured by a linear layer. We use set-to-set transformation as a more competitive module. We show that Transformer is a good choice than other implementations such as FILM~\cite{Perez2018FILM,Oreshkin2018TADAM} in Section~\ref{sec:experiment} and the supplementary.

HMS and TSP-Head change the meta-training flow, and the meta-learned embedding $\phi$ is used during meta-test.

\section{Experimental Setups}
\label{sec:experiment_setup}
We describe the concrete setups for experiments, including datasets, evaluations, and implementation details. Code is available at \url{https://github.com/hanlu-nju/revisiting-UML}.

\subsection{Datasets}
{\it Mini}ImageNet~\cite{VinyalsBLKW16Matching} contains 100 classes and 600 images per class. Three non-overlapping splits with class number 64/16/20 are used as meta-training/validation/test~\cite{Sachin2017}, respectively. {\it Tiered}ImageNet~\cite{Ren2018Meta} splits its 608 classes into 351/97/160, and there are 779,165 images in total. 
For {\it Mini}ImageNet and {\it Tiered}ImageNet, all images are cropped to $84\times84$. 
Two splits of CIFAR-100~\cite{Lee2019Meta} are investigated with 32 by 32 small images. In CIFAR-FS, 64/16/20 classes are selected from the 100 classes for meta-training/validation/test. While for FC-100, 60, 20, and 20 classes are selected in a special manner to enlarge the discrepancy among the three sets. Each class contains 600 images. No meta-training/validation labels are used by default. 

\subsection{Evaluation protocols}
Since there are no labels in the meta-validation set, we evaluate UML based on the performance of the last epoch's model.
We follow the classical protocol to evaluate the meta-learned model on few-shot classification tasks~\cite{VinyalsBLKW16Matching,SnellSZ17Prototypical,YeHZS2018Learning,Wang2019Simple}. 10,000 tasks with $N$-way $K$-shot support set are sampled from the meta-test set during the model evaluation, and there are 15 instances per $N$ classes in the corresponding query set. We set $N=5$ and $K=\{1,5,20,50\}$. We compute mean accuracy and the 95\% confidence interval. Since the confidence intervals over 10,000 trials vary around 0.15-0.20 in all settings, we omit them for clarity. Detailed results with confidence intervals are in the supplementary.

\subsection{Implementation details}
\noindent{\bf Network architectures.} We implement $\phi$ with two representative backbones. ConvNet~\cite{VinyalsBLKW16Matching,SnellSZ17Prototypical,FinnAL17Model} has four sequential blocks with convolution, batch normalization~\cite{IoffeS15BN}, ReLU, and Max Pooling. Each of the blocks outputs 64-dimensional latent embeddings, and we append a global average pooling at last. We also consider a 12-layer residual network~\cite{he2016deep,Lee2019Meta}, which is denoted as ResNet.

\noindent{\bf Optimization.} We apply Adam~\cite{KingmaB14ADAM} on ConvNet with an initial learning rate 0.002 over 100 epochs. For ResNet, we follow~\cite{Lee2019Meta,YeHZS2018Learning} and use SGD w/ momentum 0.9 over 200 epochs, whose initial learning rate is 0.03. Cosine annealing is utilized to tune the learning rate for both architectures. 

\noindent{\bf Augmentations.} We use data augmentation from AMDIM~\cite{Bachman2019Learning} by default, taking advantage of a composition of the random resized crop, random translation, color distortions, and random grayscale to construct the pseudo-labels for unsupervised meta-training.
\begin{itemize}
	\item Random resized crop makes a crop of random size (uniform from 0.08 to 1.0 in the area) of the original size and a random aspect ratio (default: of 3/4 to 4/3) of the original aspect ratio. Then the uncovered blank area is filled with reflect padding.
	\item Random translation translates vertically and horizontally by $n$ pixels where $n$ is an integer drawn uniformly and independently for each axis from $[-4, 4]$. Then the uncovered blank area is filled with reflect padding.
	\item Color distortions randomly change the brightness, contrast, and saturation of an image.
	\item Random grayscale randomly converts the image to grayscale with a probability of 0.25.
\end{itemize}

\noindent{\bf Similarity metrics.} There are four different kinds of similarity metrics, namely cosine, inner product, (negative) Euclidean distance, and our proposed SNS. 
Due to the ReLU layer at the end of the backbone, the embeddings have non-negative values.
Inner product and SNS have range $(0,+\infty)$. Euclidean ranges in $(-\infty,0]$. But the range of cosine is $[0,1]$, whose values influences the compactness of the final embeddings~\cite{Shalev2020Redesigning}. Since the best temperature varies for different configurations of tasks, we set $\tau = 0.5$ to scale the logit~\cite{chen2020simple,he2020momentum} for cosine similarity, and $\tau = 1$ for others.

\noindent{\bf Hard Mixed Supports~(HMS).} If not specified, HMS selects 10 nearest neighbors for each query. The embedding mixup coefficient is drawn from a uniform distribution, \ie, $\lambda \sim U(0,s)$. $s$ is 0.5 by default. We investigate the influence of the mixup strength $s$ in Section~\ref{sec:experiment}.

\noindent{\bf Task-Specific Projection Head~(TSP-Head).} TSP-Head uses a 1-layer and 8-head Transformer by default~\cite{Vaswani2017Attention}. The dimensionality of key, query, and value vectors is the same as the input, \ie, 64 for ConvNet and 640 for ResNet.

\section{Experimental Results}
\label{sec:experiment}
\begin{table}[tbp]
	\centering
	\caption{UML comparison on {\it Mini}ImageNet with ConvNet backbone. We report the averaged classification accuracy over 10,000 different $N$-way $K$-shot tasks.
	``Vanilla'' is the traditional UML baseline without SES and SNS.}
	\begin{tabular}{lcccc}
		\addlinespace
		\toprule
		{\bf ($N$, $K$)} & {\bf (5,1)} & {\bf (5,5)} & {\bf (5,20)} & {\bf (5,50)} \\
		\midrule
		Vanilla &43.01 & 57.94  & 65.46  & 67.62   \\
		\midrule
		CACTUs~\cite{Hsu2018Unsupervised} & 39.90  & 53.97  & 63.84  & 69.64  \\
		UMTRA~\cite{khodadadeh2019unsupervised} & 39.93  & 50.73  & 61.11  & 67.15  \\
		AAL~\cite{antoniou2019assume}   & 37.67  & 40.29  &   -    & - \\
		UFLST~\cite{ji2019unsupervised} & 33.77  & 45.03  & 53.35  & 56.72  \\
		ULDA~\cite{qin2020unsupervised}  & 40.63  & 55.41  & 63.16  & 65.20  \\
		ProtoCLR~\cite{medina2020self} & 44.89  & 63.35  & 72.27  & 74.31  \\
        CUMCA~\cite{Xu2021Unsupervised} & 41.12& 54.55& 64.45& 70.79\\
        Meta-GMVAE~\cite{Lee2021Meta} & 42.82& 55.73& 63.14& 68.26\\
		\midrule\midrule
		SimCLR~\cite{chen2020simple} & 42.09  & 55.86  & 62.17  & 63.86  \\
		MoCo-v2~\cite{chen2020improved}  & 41.97  & 55.00  & 60.59  & 62.31  \\
		MoCHi~\cite{Kalantidis2020Hard} &40.81 &55.28 & 62.89&65.11\\
		\midrule\midrule
		CACTUs {\scriptsize{\bf w/ SES+SNS}} & 44.47 & 58.24 & 65.15 & 67.08 \\
		CUMCA {\scriptsize{\bf w/ SES+SNS}} & 45.35 & 60.39   &  67.15  &  70.64    \\
		\midrule
		Baseline (Ours) & 47.43  & 64.11  & 72.52  & 74.72  \\
		HMS (Ours) & {\bf 48.12} & {\bf 65.33}  & 73.31  & 75.49  \\
		TSP-Head (Ours) & 47.35  & 65.10 & {\bf 74.45 } & {\bf 77.03 } \\
		\bottomrule
	\end{tabular}
	
	\label{tab:mini_conv}
\end{table}

We evaluate our UML baseline, HMS, and TSP-Head on FSL benchmarks. The generalization ability on novel domains and the performance change given limited base class labels are investigated. Results with Wide ResNet backbone~\cite{Zagoruyko2016Wide} and more ablation studies are in the supplementary.

\begin{table*}[t]
	\centering
	\caption{UML comparisons on {\it Mini}ImageNet, CIFAR-FS, and FC-100 with ResNet. Methods are evaluated over different $N$-way $K$-shot tasks, and we report the mean classification accuracy over 10,000 trials. $^\ast$ MoCo-v2 is also the method used in~\cite{Chen2020Shot}, but we got better results with the same backbone architecture. }
	
	\begin{tabular}{@{\;}c@{\;}cccc@{\;}||@{\;}cccc@{\;}||@{\;}cccc@{\;}}
		\addlinespace
		\toprule
		& \multicolumn{4}{c@{\;}||@{\;}}{{\it Mini}ImageNet} & \multicolumn{4}{c@{\;}||@{\;}}{CIFAR-FS} & \multicolumn{4}{c}{FC-100}\\
		\midrule
		{\bf ($N$, $K$)} & {\bf (5,1)} & {\bf (5,5)} & {\bf (5,20)} & {\bf (5,50)} & {\bf (5,1)} & {\bf (5,5)} & {\bf (5,20)} & {\bf (5,50)} & {\bf (5,1)} & {\bf (5,5)} & {\bf (5,20)} & {\bf (5,50)} \\
		\midrule
		Vanilla &   48.52  & 65.77& 73.98  & 76.20  &   46.73  & 62.93 & 67.43 & 70.76  & 32.81  & 42.17  & 51.60  & 55.02   \\
        \midrule
		SimCLR~\cite{chen2020simple} &  57.75  & 72.84  & 78.45  & 79.75 &53.86  & 69.19  & 75.22  & 77.11  & 34.69  & 47.07  & 54.87  & 57.54   \\
		MoCo-v2~$^\ast$\cite{chen2020improved}  & 54.92  & 71.18  & 77.64  & 79.30 &49.73  & 64.81  & 71.14  & 72.70  & 32.86  & 44.08  & 51.55  & 54.15 \\
		MoCHi~\cite{Kalantidis2020Hard} &57.64&75.53&82.91&84.76&48.42&63.91&71.10&72.90&35.51&46.95&54.52& 57.14\\
		\midrule
        CACTUs~\cite{Hsu2018Unsupervised}  & 55.62  & 69.50  & 74.67  & 75.68   &  51.96  & 69.08  & 74.11 & 75.38  & 35.91  & 47.29  & 53.89  & 55.94   \\
		CACTUs {\scriptsize{\bf w/ SES+SNS}} & 57.25 &	71.49 &	76.27 &	77.59 &52.53   & 73.57 & 78.48 & 79.80  & 36.79 & 48.60  & 55.52  & 57.85  \\
		\midrule
		CUMCA~\cite{Xu2021Unsupervised}  &   51.66  & 67.15 & 75.50  & 76.71  & 50.48 & 67.83  & 73.04  & 75.04  & 33.00 & 47.41  & 56.66  & 58.68     \\
		CUMCA {\scriptsize{\bf w/ SES+SNS}} & 54.95  & 71.85  & 79.06& 80.98  & 53.11 & 70.99  & 78.75  & 80.95  & 36.15  & 50.04  & 59.61  & 62.89   \\
		\midrule
		Baseline (Ours) & 56.74  & 74.05  & 81.24  & 83.04 & 53.25 & 72.05 & 80.03 & 82.16 & 37.31 & 51.62 & 61.80  & 65.54 \\
		HMS (Ours) &  {\bf 58.20} & 75.77  & 82.69  & 84.41    &     52.20  & 72.23  & {\bf 82.08} & {\bf 84.51} & {\bf 37.88} & {\bf 53.68} & {\bf 65.14} & {\bf 69.15}   \\
		TSP-Head (Ours) & 56.99  & {\bf 75.89} & {\bf 83.77} & {\bf 85.72} & {\bf 54.65} & {\bf 73.70} &  81.67 &  83.86 & 36.83 &  51.78 & 62.73 &  66.56 \\
		\bottomrule
	\end{tabular}
	
	\label{tab:mini_res}
\end{table*}

\subsection{Comparisons on Benchmarks}
The average 5-way \{1,5,20,50\}-shot classification accuracy over 10,000 trials on {\it Mini}ImageNet, {\it Tiered}ImageNet, CIFAR-FS, and FC-100 with ConvNet and ResNet backbones are listed in Table~\ref{tab:mini_conv}, Table~\ref{tab:mini_res}, and Table~\ref{tab:tiered_res}. We mark the best results in bold.
Our UML baseline generates pseudo-labels through data augmentation and utilizes sufficient episodic sampling~(SES) as well as semi-normalized similarity~(SNS). Based on this baseline, we apply our hard mixed supports (HMS) and task-specific projection head (TSP-Head).

Table~\ref{tab:mini_conv} has three blocks, including UML methods, Self-Supervised Learning~(SSL) methods, and our approaches.
``Vanilla'' denotes the traditional UML baseline using Euclidean distance and vanilla episodic sampling method. 
Some recent UML methods like UMTRA~\cite{khodadadeh2019unsupervised} using episodic training could not perform as well as the ones based on SSL methods. 
Our baseline with SNS and SES already outperforms some previous UML and SSL methods with a large margin --- for example, an 8\%-9\% accuracy superiority when compared with UMTRA~\cite{khodadadeh2019unsupervised}, and an 5\%-10\% accuracy superiority when compared with SimCLR~\cite{chen2020simple}.
ProtoCLR works similarly with ours, which learns representations in a contrastive manner over the prototypes. 
For fair comparisons, we cite the values of ProtoCLR~\cite{medina2020self} without an additional step of fine-tuning on the meta-test support set. 
Our superiority verifies that the strategy to sample pseudo tasks and the way to measure similarities are important for meta-training.
We also apply our SES and SNS together with existing UML methods (denoted as ``w/ SES+SNS'' in tables). We find SES and SNS consistently improve the vanilla baseline, as well as CACTUs~\cite{Hsu2018Unsupervised} and   CUMCA~\cite{Xu2021Unsupervised}, which demonstrate the generality of our proposed key factors in UML.

CACTUs~\cite{Hsu2018Unsupervised} is a representative clustering-based method in UML.
Once trained with our SES and SNS, CACTUs gets improved results than the reported ones in~\cite{Hsu2018Unsupervised}. CACTUs generates UML tasks based on the clustering results over the off-the-shelf learned embeddings. We try various ways to learn the embedding including those in~\cite{Hsu2018Unsupervised} and SSL methods listed in the table, then we report the best performed results. 
Although clustering may mitigate the false negative pseudo-labels in UML to some extent, we find CACTUs highly depends on the quality of the pre-learned embeddings. If pre-learned embeddings cannot differentiate instances from semantically similar classes, annotating them with different pseudo-labels during the clustering stage is hard. Another drawback of the clustering-based approaches could be the two-stage training, where the inconsistent learning objectives in the two stages degrades the performance.
We find our baseline obtains stably better results than the improved CACTUs, and the results validate that data augmentation is an effective way to generate pseudo-labels in UML.

HMS and TSP-Head achieve {\em further improvements}. In detail, HMS works well with {\em lower shots} (especially 1-shot) while TSP-Head performs better when evaluated with {\em higher shots}.\footnote{The advantages of HMS and TSP-Head show different phenomena on CIFAR variants, \eg, HMS always performs better than TSP-Head on FC-100. One possible reason is that HMS highlights the difference among smaller images via synthesized hard supports, and discriminative ability of the meta-learned embeddings helps more in this case.} 
As reported by~\cite{Chen2020Shot}, MoCo achieves 56.2\% and 75.4\% classification accuracy on 1-shot and 5-shot {\it Mini}ImageNet tasks upon the ResNet-50 backbone, respectively. Compared with our results in Table~\ref{tab:mini_res}, our proposed UML methods get better performance with the shallow ResNet-12 backbone.

We make detailed comparisons between HMS and MoCHi~\cite{Kalantidis2020Hard}, where MoCHi uses synthesized hard negatives in contrastive SSL. 
We find HMS performs better in most cases without using a memory bank. The success of HMS may come from the special configuration of episodic tasks during meta-training. We provide the detailed influence of SES on HMS in the supplementary.

\begin{table}[t]
	\centering
	\caption{The mean classification accuracy on {\it Tiered}ImageNet with ResNet backbone. Methods are evaluated over 10,000 $N$-way $K$-shot tasks.}
	\begin{tabular}{lcccc}
		\addlinespace
		\toprule
{\bf ($N$, $K$)} & \textbf{(5,1) }                & \textbf{(5,5) }                & \textbf{(5,20) }               & \textbf{(5,50) }               \\
\midrule
SimCLR~\cite{chen2020simple}          & 57.87            & 73.74          & 79.71        & 81.18            \\
MoCo-v2~\cite{chen2020improved}         & 56.63          & 74.98            & 82.14            & 83.98           \\
MoCHi~\cite{Kalantidis2020Hard}           & 58.40         & 75.49        & 82.08          & 83.87            \\
\midrule
CACTUs {\scriptsize{\bf w/ SES+SNS}} & 56.47          & 71.09         & 76.42         & 77.66 \\
Baseline (Ours) & 56.29        & 74.11      & 81.64       & 83.70        \\
HMS (Ours)      & \textbf{58.42} & \textbf{75.85} & \textbf{82.58} & 84.24         \\
TSP-Head (Ours) & 56.46       & 74.85        & 82.56         & \textbf{84.52} \\
		\bottomrule
	\end{tabular}
	\label{tab:tiered_res}
\end{table}

\begin{figure}[t]
	\centering
	\includegraphics[width=1.0\linewidth]{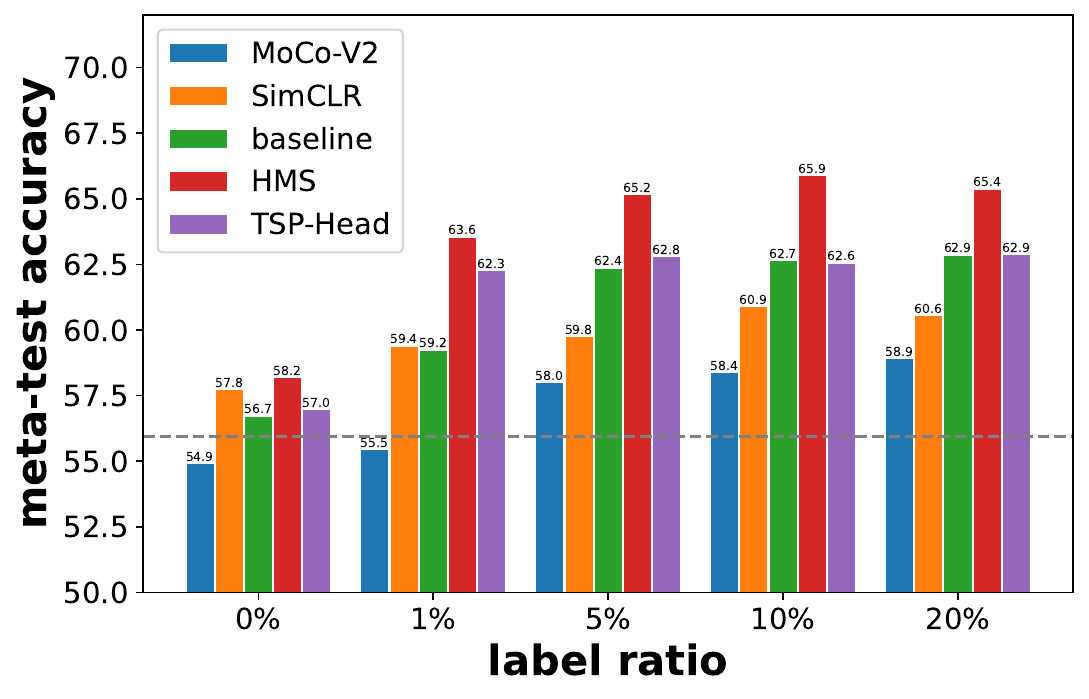}
	\caption{The change of 5-way 1-shot classification accuracy on {\it Mini}ImageNet with ResNet when the label ratio in the base class set increases. The dotted line is the performance of the ProtoNet learned with a fully labeled meta-training set.}\label{fig:fine-tune-res}
	
\end{figure}

\noindent{\bf Limited base class labels lead to a strong FSL model.} 
We investigate how the embedding learned with UML facilitates supervised meta-learning given limited base class labels.
Given the initialized embedding $\phi$, we fine-tune the same objective in Eq.~\ref{eq:meta_obj}, but the episodes of tasks are constructed based on the ground-truth labels of base class examples. Our proposed SES and SNS are applied to make the two-stage embedding updates consistent.

We fine-tune the embedding for 50 and 100 epochs when the labeling ratio is below and above 10\%, respectively. There are ten episodes per epoch. In each episode, we sample a mini-batch consisting of 256 images from 64 base classes with four images in each class. Based on the sampled batch, 512 tasks are further re-sampled via SES. The learning rate is 0.0001, and the optimizer is SGD w/ momentum. 

Results based on ResNet are shown in Fig~\ref{fig:fine-tune-res}. 
We control the fraction of labeled instances in each base class. The zero label ratio corresponds to the fully unsupervised case (the same as UML), which is then utilized as the initial weights for other analyses. 
By fine-tuning $\phi$ with more labeled base class instances, FSL accuracy on the meta-test set increases due to richer supervisions and even exceeds the fully supervised ProtoNet meta-trained on a labeled base class set from scratch~\cite{SnellSZ17Prototypical} (the dotted line in the figure). The phenomenon indicates that the UML pre-trained weights are tremendously label-efficient --- we get a strong FSL model given a large unlabeled base class set and a small number of labels.

Some semi-supervised meta-learning methods utilize the unlabeled data in meta-training set~\cite{Ren2018Meta,Ye2021TACO}, but could not perform as well as the fully supervised upper bound.
Surprisingly, we find that based on the UML pre-trained weights and with only 1\% labels, fine-tuning via classical meta-learning method can achieve performance comparable to a fully supervised model. When pre-trained with HMS, the fine-tuned model with 10\% labels can even be competitive to a number of state-of-the-art supervised FSL methods using supervised pre-trained weights with full labels.

Similar observations are also verified in~\cite{Chen2020Big} that semi-supervised classification improves a lot with self-supervised learned weights. 
We verify the importance of the unsupervised learned weights to supervised meta-learning. 
We also initialize the embedding with weights learned by SSL methods in the same FSL configuration. We find results based on MoCo and SimCLR improve with more labels but are far away from those based on UML-learned embeddings given more labels. 
One possible reason is the inconsistency between the pre-training and fine-tuning objectives.

\begin{table}[t]
	\centering
	\caption{Mean classification accuracy of methods learned from the base class set of {\it Mini}ImageNet and evaluated on $N$-way $K$-shot novel class tasks of CUB.}
	
	\begin{tabular}{cccc}
		\addlinespace
		\toprule
		{\bf ($N$, $K$)} & {\bf (5,1)} & {\bf (5,5)} & {\bf (5,20)} \\
		\midrule
		SimCLR~\cite{chen2020simple} & 38.91 & 54.88 & 63.76 \\
		MoCo-v2~\cite{chen2020improved}  & 37.52 & 51.04 & 59.48 \\
		\midrule
		Baseline (Ours) & 39.72 & 55.45 & 64.45 \\
		HMS (Ours) & 40.65 & 57.56 & 67.57 \\
		TSP-Head (Ours)  & {\bf 40.75} & {\bf 58.32} & {\bf 68.61} \\
		\bottomrule
	\end{tabular}
	\label{tab:cub}
	
\end{table}

\subsection{Cross-Domain Comparisons}
Since the embeddings are learned in an unsupervised manner, it could be memoryless of the base-class information and become more generalizable on novel domains. With the embedding meta-learned on the base class set of {\it Mini}ImageNet, we evaluate (meta-test) it on CUB~\cite{WahCUB_200_2011}. CUB is a fine-grained dataset on different species of birds. Following the splits of~\cite{YeHZS2018Learning}, 50 classes are randomly selected from CUB as the meta-test set, where 5-way \{1,5,20\}-shot tasks are sampled for embedding evaluation in Table~\ref{tab:cub}. 

Results verify that improvements on the in-domain evaluations also generalize to novel domains. We find TSP-Head gets the best results in all cases, which is consistent with its motivation --- an auxiliary task-specific projection head makes the pre-adapted embedding becomes more generalizable even to other domains.

\subsection{Ablation Studies}
We further investigate the properties of the UML learned embedding on {\it Mini}ImageNet with ResNet backbone.

\begin{table}[tbp]
	\centering
	\caption{Mean classification accuracy evaluated on different $N$-way $K$-shot tasks on {\it Mini}ImageNet. The top and bottom parts are unsupervised and supervised methods, respectively. $\dagger$ denotes the method that utilizes initialization weights pre-trained over all labeled base classes.}
	\begin{tabular}{ccccc}
		\addlinespace
		\toprule
		{\bf ($N$, $K$)} & {\bf (5,1)} & {\bf (5,5)} & {\bf (5,20)} & {\bf (5,50)}\\
		\midrule
		Baseline (Ours) & 56.74  & 74.05  & 81.24  & 83.04 \\
		HMS (Ours) &  58.20 & 75.77 & 82.69 & 84.41\\
		TSP-Head (Ours)  & 56.99  &  75.89  & {\bf 83.77 } & {\bf 85.72 } \\
		\midrule
		ProtoNet~\cite{SnellSZ17Prototypical} & 55.93 & 69.68 & 74.82 & 76.17 \\
		ProtoNet$^\dagger$ & {\bf 63.09} &{\bf 78.15} & 83.19 & 84.41 \\
		\bottomrule
	\end{tabular}
	\label{tab:upper-bound}
\end{table}
\begin{table}[t]
	\centering
	\caption{Comparison between HMS and other mixup-based methods.  
    We record the $N$-way $K$-shot classification accuracy on the meta-test set of {\it Mini}ImageNet with ResNet.}
	\begin{tabular}{ccccc}
		\addlinespace
		\toprule
		{\bf ($N$, $K$)} & {\bf (5,1)} & {\bf (5,5)} & {\bf (5,20)} & {\bf (5,50)}\\
		\midrule
		MetaMix~\cite{Yao21Improving} & 50.23 & 68.30  & 71.02 & 73.62  \\
		CUMCA~\cite{Xu2021Unsupervised} &  51.66  & 67.15 & 75.50  & 76.71    \\
		Hard Task~\cite{Sun2019Meta,Sun2020Meta} &54.74 & 72.61 & 78.96  & 80.93 \\
		\midrule
		HMS & {\bf 58.20} & 75.77  & 82.69  & 84.41  \\
		\bottomrule
	\end{tabular}
	\label{tab:mixup_compare}
\end{table}
\begin{table}[tbp]
	\centering
	\caption{The influence of mixup strength in HMS. All experiments select 10 nearest neighbors for each query instance but mixed up with coefficients sampled from different uniform distributions. We record the $N$-way $K$-shot classification accuracy on  the meta-test set of {\it Mini}ImageNet with ResNet backbone. When $\lambda = 0$, HMS only augments the support set with the nearest neighbors with different pseudo-classes without performing mixup.}
	\begin{tabular}{lcccc}
		\addlinespace
		\toprule
		{\bf ($N$, $K$)} & {\bf (5,1)} & {\bf (5,5)} & {\bf (5,20)} & {\bf (5,50)} \\
		\midrule
		baseline (w/o HMS) &56.74  & 74.05  & 81.24  & 83.04  \\
		$\lambda = 0$ & 55.86  & 75.02 & 83.41 & 85.43   \\
		$\lambda \sim U(0,0.05)$ & 56.37 & 75.49 & 83.57 & 85.63  \\
		$\lambda \sim U(0,0.1)$ & 56.55 & 75.60 & \textbf{83.71} & \textbf{85.66} \\
		$\lambda \sim U(0,0.2)$ & 57.03 & 75.66 & 83.40 & 85.19  \\
		$\lambda \sim U(0,0.5)$ & 58.20 & \textbf{75.77} & 82.69 & 84.41  \\
		$\lambda \sim U(0,1.0)$ & {\bf 59.27} & 74.55 & 80.46 & 81.82 \\
		\bottomrule
	\end{tabular}
	\label{tab:mixup}
\end{table}

\noindent{\bf Comparison with the supervised upper-bound.}
The embedding learned without base class labels show strong discriminative ability on benchmarks. We compare UML embedding with its supervised upper bound (ProtoNet~\cite{SnellSZ17Prototypical}) in Table~\ref{tab:upper-bound}.
There are two versions of supervised upper-bound. One optimizes the meta-learning objective in Eq.~\ref{eq:meta_obj} from scratch directly, and another one fine-tunes the meta-learning objective with supervised pre-trained weights. We denote the latter one with $\dagger$ in the table. Since the supervised pre-training optimizes a classifier over all base classes, it improves the meta-learned embeddings a lot~\cite{YeHZS2018Learning}. We find our UML variants outperform the vanilla ProtoNet in all cases. The TSP-Head even gets better classification accuracy than the supervised pre-trained ProtoNet when meta-tested with higher-shots. The results indicate the strong discriminative and generalization ability of the UML embedding.

\noindent{\bf Comparison with other mixup strategies.}
HMS creates query-specific hard tasks by mixing up the mined confusing instances with a query instance.
The same notion is also applied in various supervised meta-learning methods. For example, HT~\cite{Sun2019Meta,Sun2020Meta} synthesizes hard tasks with selected confusing classes in advance. MetaMix~\cite{Yao21Improving} mixups embeddings between support and query sets. CUMCA~\cite{Xu2021Unsupervised} proposes Prior-Mixup as a kind of data augmentation for UML.
We compare HMS with these methods in Table~\ref{tab:mixup_compare}. 
The results indicate that HMS constructs confusing tasks in a more effective manner.

\noindent{\bf The strength to mixup embeddings in HMS.}
By default, the mixup coefficient is sampled from a uniform distribution ranging from 0 to 0.5, \ie, $\lambda \sim U(0, 0.5)$. We set the upper bound as 0.5 to ensure the mixed embedding is biased towards the mined instance and semantically different from the query~\cite{Kalantidis2020Hard}. Although HMS with the default range show promising results, we investigate the influence of mixup coefficient strength, \ie, sampling $\lambda \sim U(0,s)$ with different $s$, \eg, $s=0.05$ or even $s=1.0$.

Table~\ref{tab:mixup} shows that when the mixup strength $s$ becomes larger, HMS gets better few-shot classification accuracy, especially 1-shot and 5-shot. Different from~\cite{Kalantidis2020Hard}, our HMS achieves the best 1-shot classification performance with $s=1.0$, which we could not preserve the semantic meaning of the synthesized hard support instance. One possible reason is that in UML we do not have a strict requirement on the negative candidates in a pseudo task.
Therefore, {\em larger mixup strength helps HMS with lower shots and smaller strength facilitates higher shots.}
We keep $s=0.5$ in all our experiments, but with carefully selected $s$, HMS is able to get better results especially in 1-shot, 20-shot, and 50-shot tasks as Table~\ref{tab:mixup}.

\begin{table}[tbp]
	\centering
	\caption{We evaluate the pre-adapted (w/o head) and post-adapted (w/ head) embeddings on two kinds of tasks based on the learned  TSP-Head. The ``pseudo tasks'' are generated based on pseudo-classes determined by data augmentations as in meta-training, while the ``real tasks'' are sample based on real semantic classes. 
	Methods are evaluated over $N$-way $K$-shot tasks on {\it Mini}ImageNet with ResNet.}
	\begin{tabular}{lcccc}
		\addlinespace
		\toprule
{\bf ($N$, $K$)} & \textbf{(5,1)}                & \textbf{(5,5)}                & \textbf{(5,20)}               & \textbf{(5,50)}               \\
\midrule
\multicolumn{5}{l}{\it Meta-test on pseudo tasks} \\ 
w/ head                & \textbf{98.87} & \textbf{99.73} & \textbf{99.78} & \textbf{99.80} \\
w/o head               & 98.64         & 99.72          & 99.77          & 99.79          \\
\midrule
\multicolumn{5}{l}{\it Meta-test with real tasks} \\
w/ head                & 56.90          & 75.87          & 83.59         & 85.69         \\
w/o head               & \textbf{56.99} & \textbf{75.94} & \textbf{83.72} & \textbf{85.74} \\
		\bottomrule
	\end{tabular}
	\label{tab:task_dist}
		\vspace{-4pt}
\end{table}

\begin{table}[tbp]
	\centering
	\caption{Mean classification accuracy between task-agnostic projection head / task-specific projection head on different $N$-way $K$-shot tasks on {\it Mini}ImageNet. 
	In addition to Transformer, we also investigate another implementation of the TSP-Head based on DeepSets and FiLM.}
	
	\begin{tabular}{ccccc}
		\addlinespace
		\toprule
		{\bf ($N$, $K$)} & {\bf (5,1)} & {\bf (5,5)} & {\bf (5,20)} & {\bf (5,50)}\\
		\midrule
		\multicolumn{5}{l}{\it Task-agnostic projection head} \\
		Projection Head~\cite{chen2020simple} & 53.32 & 70.87 & 78.38 & 80.32\\
		\midrule
\multicolumn{5}{l}{\it Task-specific projection head} \\
		DeepSets~\cite{Zaheer17Deep}+FiLM~\cite{Perez2018FILM} & 51.71 & 73.28 & 82.15 &84.48 \\
		TSP-Head (ours) & \bf 56.99  & \bf 75.89 & \bf 83.77 & \bf 85.72 \\
		\bottomrule
	\end{tabular}
	\label{tab:projection-head}
	\vspace{-4pt}
\end{table}

\noindent{\bf Does TSP-Head bridge the task distribution gap?} 
As mentioned in Section~\ref{sec:tsp_head}, one main motivation of TSP-Head is to mitigate the negative effects of the task distribution gap between the pseudo-labeled tasks in meta-training and the real-labeled tasks in meta-test.
We evaluate the learned TSP-Head on two types of tasks on the meta-test set -- ``pseudo tasks'' sampled based on pseudo-classes (determined by data augmentations as in meta-training), and ``real tasks'' that are sampled based on real semantic classes. 
Given a learned TSP-Head model, we report the accuracy w/ and w/o the Transformer head, \ie, using the post-adapted $\psi(\x)$ and pre-adapted $\phi(\x)$ in Eq.~\ref{eq:trans}, respectively. 
Since TSP-Head is applied over both support and query sets during meta-training, we keep the same in meta-test when using the Transformer head. In other words, ``w/ head'' adapts instances in a transductive way. More discussions are in the supplementary.
From Table~\ref{tab:task_dist}, we find when evaluating on pseudo tasks, the same task distribution as meta-training, maintaining the projection head and using the post-adapted embedding perform better. In contrast, using the pre-adapted embedding without the projection head generalizes better on real tasks (the same as our UML evaluations), which verifies that TSP-Head eases the problem of overfitting pseudo tasks. 

\noindent{\bf Is task-specific projection head necessary?}
The projection head is a useful component in SSL, \eg, SimCLR~\cite{chen2020simple}, where a nonlinear projection layer is appended over the top-layer embedding. We also implement the projection head as a task-agnostic one, and the results are in Table~\ref{tab:projection-head}. The advantage of the TSP-Head indicates that the explicit consideration of the specific properties of sampled tasks during meta-training is necessary. We also compare our Transformer-based task-specific projection head with another implementation -- DeepSets~\cite{Zaheer17Deep} + FiLM~\cite{Perez2018FILM}. In this method, we use DeepSets to generate task-specific linear transformation parameters for the FiLM layer. Detailed configurations are in the supplementary. Table~\ref{tab:projection-head} shows that when compared with the task-agnostic one, TSP-Head generally improves UML when evaluated on higher shot tasks. Among the implementations, our Transformer-based TSP-Head is the best choice.

\begin{table}[t]
	\centering
	\caption{Mean accuracy between HMS, TSP-Head, and their combination on $N$-way $K$-shot tasks on {\it Mini}ImageNet.}
	
	\begin{tabular}{ccccc}
		\addlinespace
		\toprule
		{\bf ($N$, $K$)} & {\bf (5,1)} & {\bf (5,5)} & {\bf (5,20)} & {\bf (5,50)}\\
		\midrule
		HMS & {\bf 58.20 } & 75.77  & 82.69  & 84.41  \\
		TSP-Head & 56.99  & {\bf 75.89} & {\bf 83.77} & {\bf 85.72} \\
		TSP-Head + HMS & 56.74 & 74.58 & 82.21 & 84.11\\
		\bottomrule
	\end{tabular}
	\label{tab:combine}
	
\end{table}

\noindent{\bf Combine two techniques together?}
We propose two methods to take the characteristics of tasks into account. An intuitive question is whether the two methods could be fused to get further improvements. As shown in Table~\ref{tab:combine}, we find HMS and TSP-Head are not compatible, and directly combining them cannot get better results. One main reason could be that HMS and TSP-Head improve the embeddings from two different perspectives, so we need some special strategies to combine their advantages together.

\subsection{Linear Evaluation}
\begin{figure}[t]
	\centering
	\includegraphics[width=0.9\linewidth]{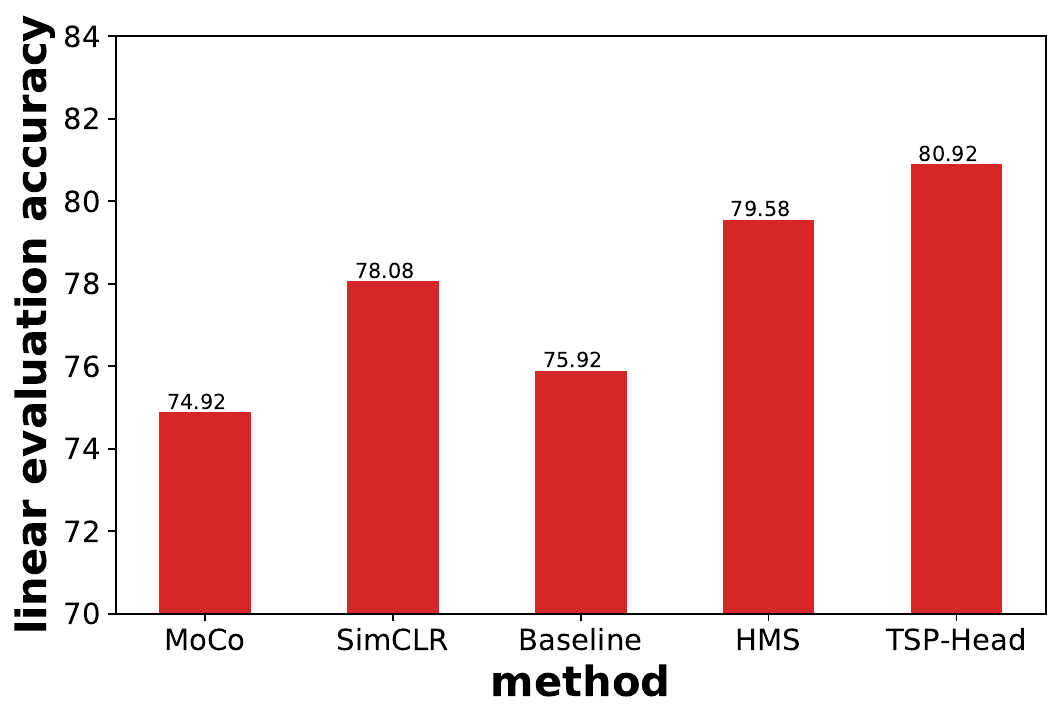}
	\caption{Linear evaluation comparisons. The meta-test set is split into 10 folds. A logistic regression is trained on frozen embeddings of the 9 folds, and the remaining fold is used to evaluate accuracy. All methods are learned on the meta-training set of {\it Mini}ImageNet with ResNet.}
	\label{fig:linear-eval}
\end{figure}

We also evaluate our UML methods following the SSL protocol. In detail, we \textit{freeze} the feature extractor (\ie, the embedding) and train a linear classifier on top of it. After extracting the features of all instances in the meta-test set, we randomly split the meta-test set into 10 folds. Each fold has the same number of samples from each class. 9 of the 10 folds are used to train a linear logistic regression, and we test the learned linear classifier on the remaining one fold.
The logistic regression is trained with $\ell_2$ regularization, whose weight is searched in a logarithmic scale between $10^{-4}$ and $10^4$ with 5-fold cross-validation. We report test accuracy in Fig.~\ref{fig:linear-eval}. Two representative SSL methods, MoCo-v2~\cite{chen2020improved} and SimCLR~\cite{chen2020simple}, are compared, whose hyper-parameters are consistent with those in their published papers. 

\begin{figure*}[t]
	\centering
	\subfigure{
	\hspace{1mm}
	\includegraphics[width=0.6\linewidth]{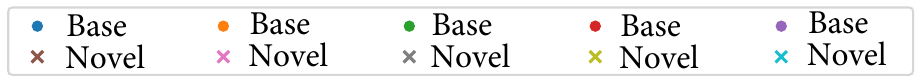}
	}\\
	\setcounter{subfigure}{0}
	\subfigure[ProtoNet (supervised)]{
		\label{fig:subfig:protonet} 
		\includegraphics[scale=0.3]{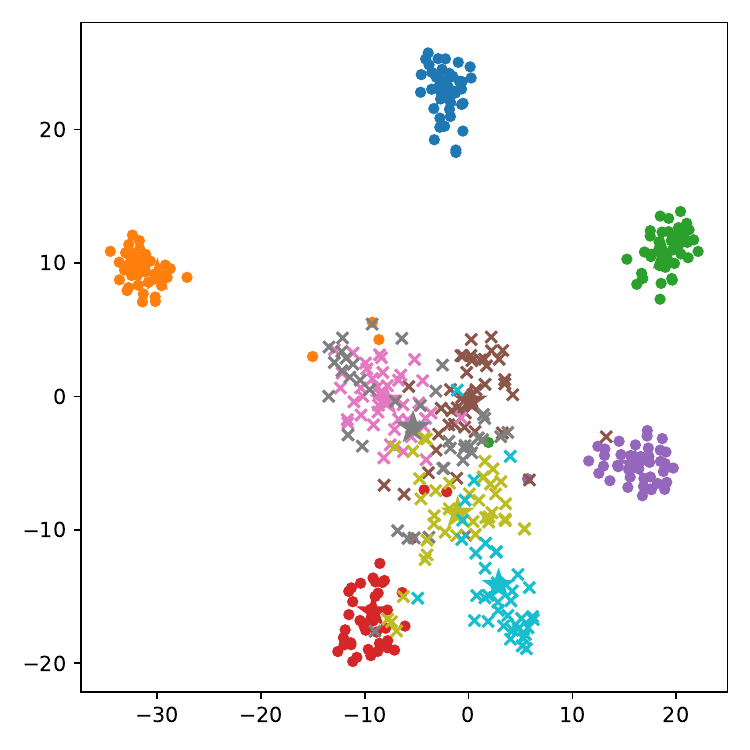}
	}
	\subfigure[baseline (ours)]{
		\label{fig:subfig:baseline} 
		\includegraphics[scale=0.3]{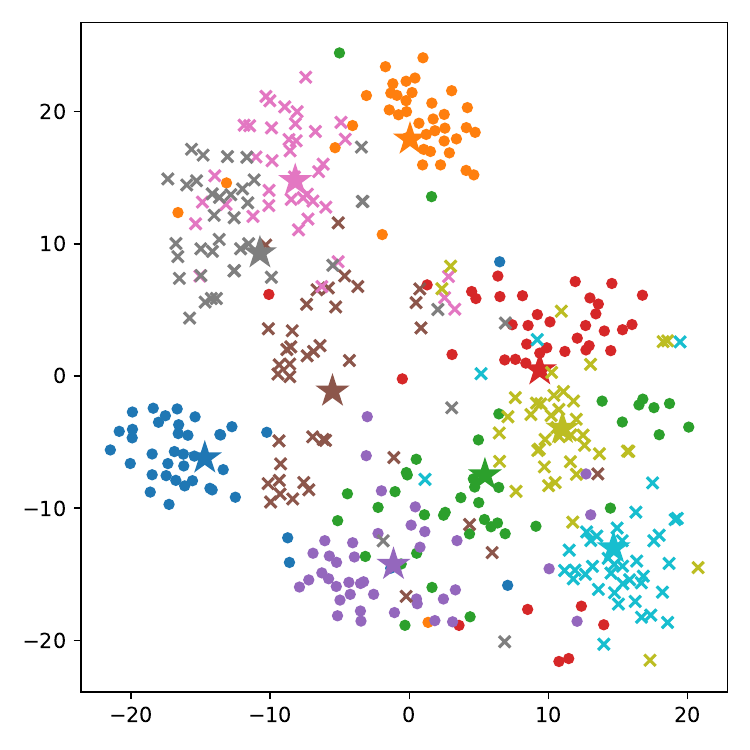}
	}
	\subfigure[HMS (ours)]{
		\label{fig:subfig:hms} 
		\includegraphics[scale=0.3]{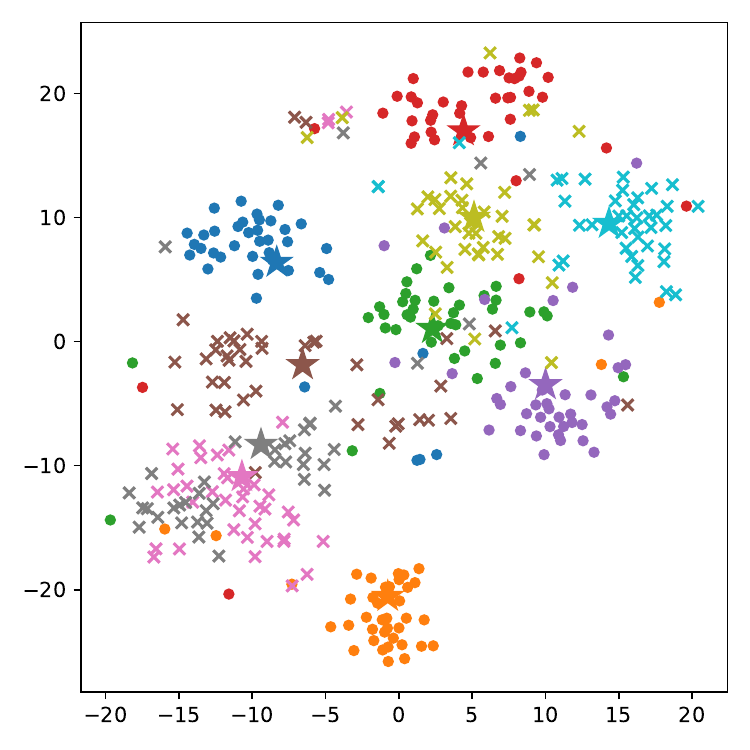}
	}
	\subfigure[TSP-Head (ours)]{
		\label{fig:subfig:tsp} 
		\includegraphics[scale=0.3]{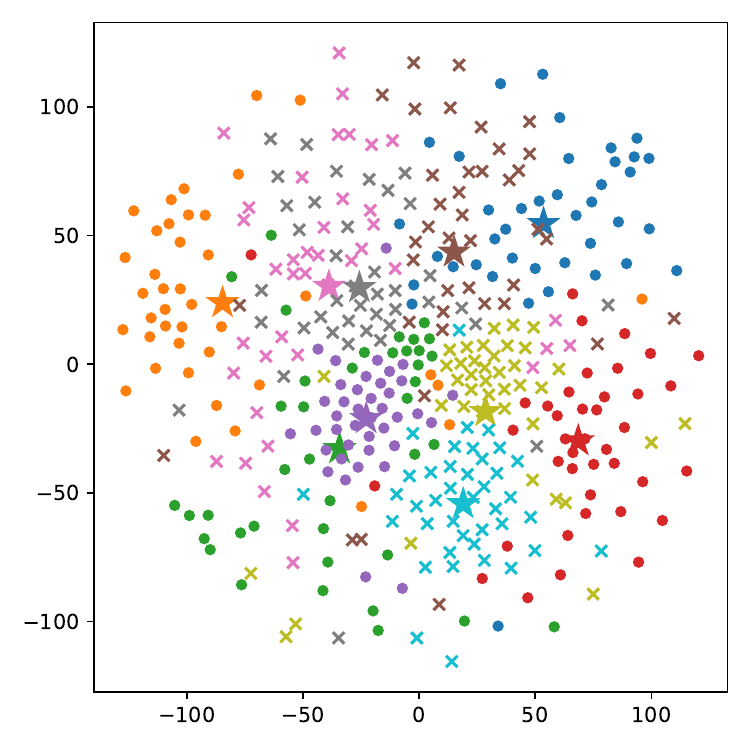}
	}
	\caption{T-SNE visualization of learned embeddings on the {\it Mini}ImageNet. Four plots display results of ProtoNet~\cite{SnellSZ17Prototypical}~(a), UML baseline~(b), HMS~(c), and TSP-Head~(d). We use circle and cross to represent ``base'' and ``novel'' classes, respectively. ProtoNet is trained in a supervised manner, and the other three are trained without using any base class labels.}
	\label{fig:twopicture} 
\end{figure*}

With this special evaluation protocol, we find although our UML baseline outperforms SSL methods on few-shot learning tasks, it does not necessarily have an advantage on other downstream tasks like linear classification.
By equipping the baseline with HMS or TSP-head, it achieves better results as shown in Fig.~\ref{fig:linear-eval}, which indicates that through incorporating characteristics of tasks, the learned embeddings can be more generalizable to handle such a heterogeneous linear evaluation task.
The TSP-Head achieves the best result, which is consistent with our observation that TSP-Head facilitates more shot tasks. We hope the evaluations with few-shot tasks and with linear models will draw more insights on both the UML and the SSL.

\subsection{Visualize learned embeddings}
In Fig.~\ref{fig:twopicture}, we visualize the learned embedding with t-SNE and show the difference when the embedding is applied over meta-training (denoted as ``base'') and meta-test (denoted as ``novel'') sets of {\it Mini}ImageNet.
In detail, we randomly select 5 classes with 40 instances per class from both splits. 
Different colors represent different classes. "$\star$" marks the center of the corresponding class.

As shown in~\ref{fig:subfig:protonet}, the embeddings learned in a supervised manner present a good clustering effect on base classes, but the embeddings of novel classes are close to each other and can not be easily separated. 
On the contrary, our UML methods show similar properties on both base and novel classes, in either \ref{fig:subfig:baseline} or \ref{fig:subfig:hms}. They form kinds of clusters but do not overfit on base class data, which indicates that the embeddings learned with UML may have more generalization potential. Fig.~\ref{fig:subfig:tsp} shows obviously different embeddings learned by TSP-Head. Samples are scattered among the space, keeping a certain distance from each other. This kind of uniformity may explain why TSP-Head is good at 20-shot and 50-shot tasks.

\section{Conclusion}
Instead of learning to learn embeddings for few-shot classification through a labeled base class set, we propose to transform the meta-learning methods to a fully {\em unsupervised} manner from both sampling and modeling aspects {\em without noticeable performance degrades}.
Simple modifications like Sufficient Episodic Sampling~(SES) and Semi-Normalized Similarity (SNS) lead to strong Unsupervised Meta-Learning~(UML) baselines. 
We further take full advantage of the characteristic of tasks from two directions. We consider Hard Mixed Supports~(HMS) constructing difficult meta-training tasks dynamically. Moreover, another strategy utilizing Task-Specific Projection Head~(TSP-Head) takes the divergence across tasks into account. Both approaches make the meta-learned embeddings more discriminative and generalizable.
Our proposed UML methods outperform other UML models on few-shot classification benchmarks. They also achieve similar state-of-the-art performance with its supervised variants given only 10\% of base class labels. 

\ifCLASSOPTIONcompsoc
  \section*{Acknowledgments}
\else
  \section*{Acknowledgment}
\fi

Thanks to Hexiang Hu for valuable discussions. This work is partially supported by The National Key
R\&D Program of China (2020AAA0109401), NSFC (62006112, 61773198, 61921006), NSF of Jiangsu Province (BK20200313), CCF-Baidu Open Fund (NO.2021PP15002000).

{\small
\bibliographystyle{IEEEtran}
\bibliography{references}}

 \newpage

\appendices

\section{Tasks Configurations for UML}
\label{sec:appendix_analysis}
\begin{figure}
	\begin{minipage}{1\linewidth}
		\centering
		\includegraphics[width=\textwidth]{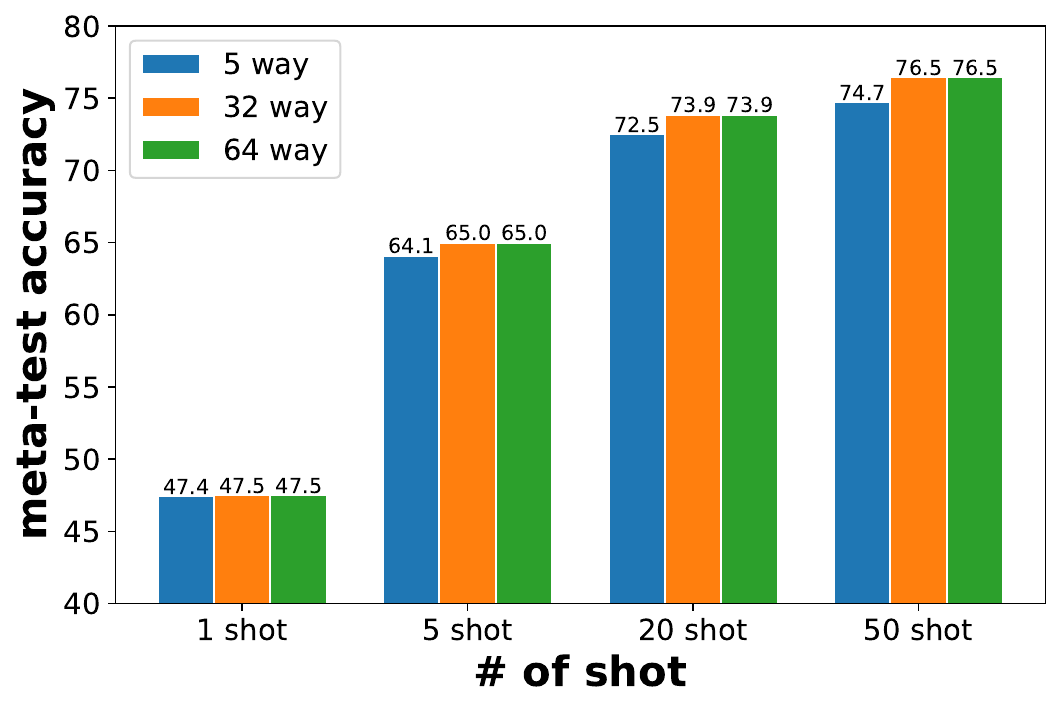}
	\end{minipage}
	\caption{Investigation on the number of way $N\in\{5,32,64\}$ during meta-training for UML. All meta-learned embeddings  are evaluated on $5$-way $\{1,5,20,50\}$-shot classification tasks over 10,000 trials on {\it Mini}ImageNet. Please see the texts for more details.
	We observe {\em by increasing the number of ways we can get more discriminative embeddings on more shot evaluations, and further increase of the way value $N$ does not show apparent contribution when $N$ exceeds some threshold}.}\label{fig:way_bar}
\end{figure}

\begin{figure}
	\begin{minipage}{1\linewidth}
		\centering
		\includegraphics[width=\textwidth]{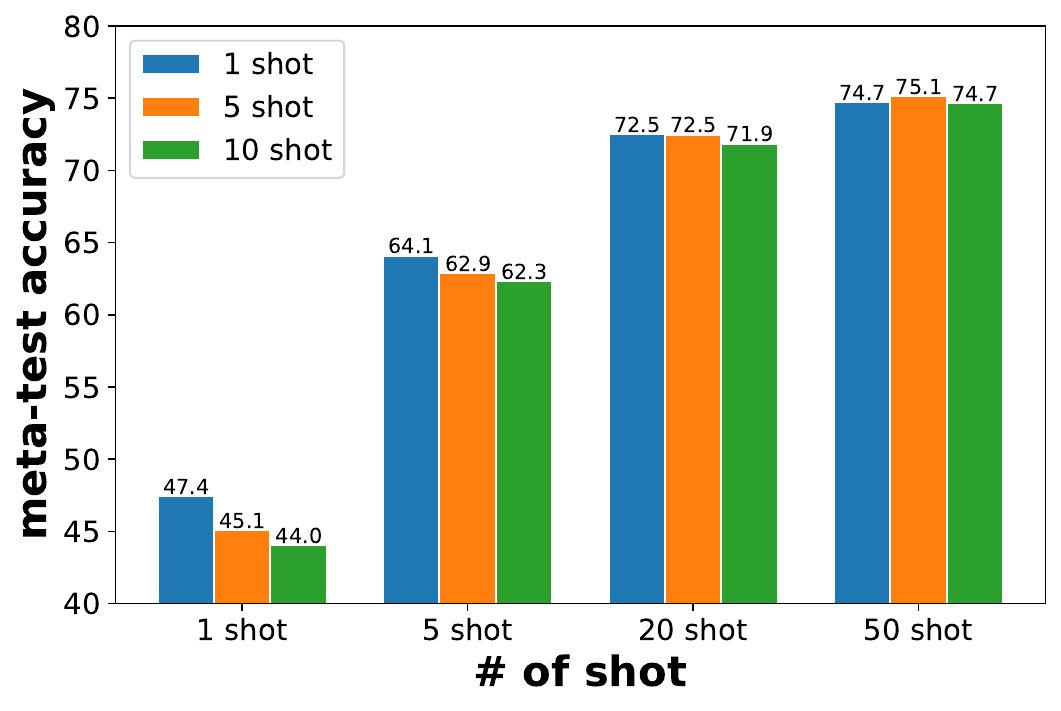}
	\end{minipage}
	\caption{Investigation on the influence of the support set shot number $K\in\{1,5,10\}$ in a task. 
	All meta-learned embeddings are evaluated on $5$-way $\{1,5,20,50\}$-shot classification tasks over 10,000 trials on {\it Mini}ImageNet. Please see the texts for more details.
	We find {\em using smaller shot numbers in a task generalizes better, especially in the 1-shot meta-test case.}
	}\label{fig:shot_bar}
\end{figure}

\begin{figure}
	\begin{minipage}{1\linewidth}
		\centering
		\includegraphics[width=\textwidth]{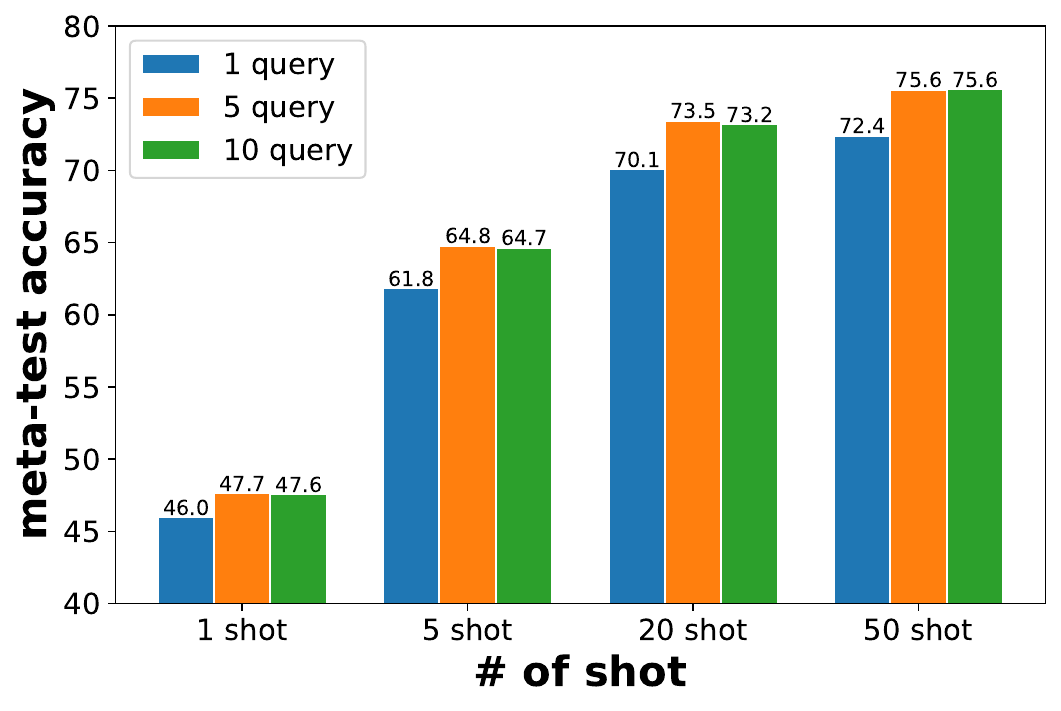}
	\end{minipage}
	\caption{Investigation on the number of shot in the query set $Q\in\{1,5,10\}$ during meta-training of UML. 
	All meta-learned embeddings are evaluated on $5$-way $\{1,5,20,50\}$-shot classification tasks over 10,000 trials on {\it Mini}ImageNet. Please see the texts for more details.
	The figure illustrates that {\em the number of query instances needs to reach a certain value} to make the meta-learned embeddings generalizable.}\label{fig:query_bar}
\end{figure}

\begin{figure}
	\centering
	\includegraphics[width=\textwidth]{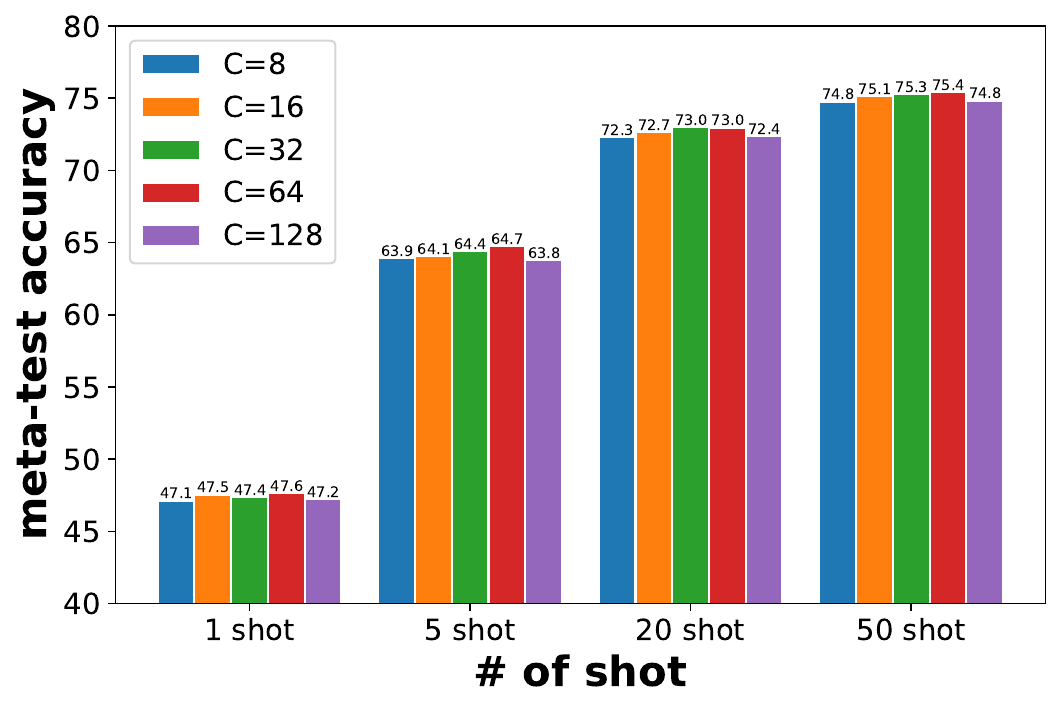}
	\caption{Investigation on the number of sampled instances $C$ per episode during meta-training of UML. The larger the value of $C$, the larger the size of the mini-batch in meta-training.
	All meta-learned embeddings are evaluated on $5$-way $\{1,5,20,50\}$-shot classification tasks over 10,000 trials on {\it Mini}ImageNet. Please see the texts for more details.
	The figure illustrates that {\em our UML baseline is insensitive to the mini-batch size}.}
	\label{fig:batch_bar}
\end{figure}

\begin{figure}
	\begin{minipage}{1\linewidth}
		\centering
		\includegraphics[width=\textwidth]{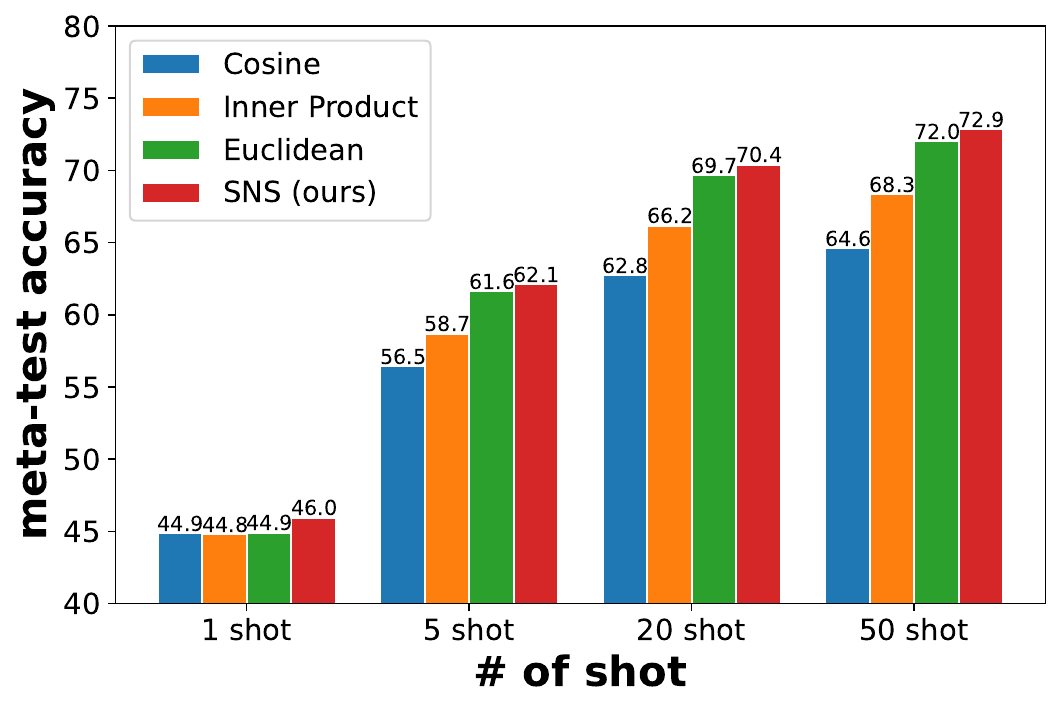}
	\end{minipage}
	\caption{Comparison of different similarities for UML. We sample one task per episode (the task number is set to one in SES) during meta-training. 
	All meta-learned embeddings  are evaluated on $5$-way $\{1,5,20,50\}$-shot classification tasks over 10,000 trials on {\it Mini}ImageNet with ConvNet backbone. Please see the texts for more details.
    The figure illustrates {\em our SNS is superior to other existing similarities}.}\label{fig:similarity}
\end{figure}

As mentioned in Section 4 in the main paper, there are various key factors for (pseudo) task configurations during meta-training. We empirically explore how they influence the Unsupervised Meta-Learning (UML) with our UML baseline and find the best configuration of tasks.

Following the same setting in Section 4, we implement the feature extractor $\phi$ with a four-layer ConvNet and focus on the {\it Mini}ImageNet~\cite{VinyalsBLKW16Matching} benchmark with the standard splits. During meta-training, we optimize over episodes of (pseudo) tasks over 100 epochs. 
In each episode, we sample $C$ instances and make $K+Q$ random augmentations for each of them. So the batch-size is $C\times(K+Q)$ with $C$ different pseudo-classes.\footnote{We enumerate all base class instances per epoch with multiple episodes, and a larger $C$ decreases the number of episodes per epoch.}
We sample 512 $N$-way $K$-shot support sets corresponding with $N$-way $Q$-shot query sets from the mini-batch with Sufficient Episodic Sampling (SES).
Semi-Normalized Similarity is applied in the embedding-based classifier in a (pseudo) task by default, which measures whether two instances are similar or not.
The initial learning rate is set to 0.002 and is cosine annealed. 
The {\em default values} of the factors $N$, $K$, $Q$, and $C$ are $5$, $1$, $5$, $64$, respectively.
We investigate the influence of $N$, $K$, $Q$, and $C$ in meta-training by varying a particular factor and keeping others as those default values.

During meta-test, we evaluate the learned embedding $\phi$ over 10,000 5-way $K=\{1,5,20,50\}$-shot tasks, which means we change the shot number in the support set, and we keep $Q=15$ for all query sets.
We only use labels to evaluate some statistics such as few-shot classification accuracy, and {\em no base class labels are utilized during meta-training}.

\noindent{\bf The influence of the ``way'' number $N$.} In Fig.~\ref{fig:way_bar}, we fix $K = 1, Q = 5$ and vary $N\in\{5,32,64\}$. 
We observe {\em by increasing the number of ways $N$, we can get more discriminative embeddings on more shot evaluations, and further increase of $N$ does not show apparent contribution when $N$ exceeds some threshold}. 
One explanation for this phenomenon is that, similar to the supervised scenario~\cite{SnellSZ17Prototypical}, the number of ways is related to the difficulty of the sampled (pseudo) tasks during meta-training. The more difficult the meta-training process is, the better model we have. 
In other words, we sample $N$-way $K$-shot tasks from the mini-batch {\em by increasing $N$ as large as $C$}. 
Although larger $N$ (as well as larger $C$ since $C=N$) decreases the number of shots and query instances in a task when we fix the size of mini-batch, the larger-way tasks promote the diversity of instances and facilitate UML. 

\noindent{\bf The influence of the ``shot'' number $K$.} Since we observe sampling larger-way tasks always achieves better results, so we fix $N=C=64$ in Fig.~\ref{fig:shot_bar}. 
By fixing $Q=5$ during meta-training, we change the number of shots $K=\{1,5,10\}$. It is notable that by increasing $K$ those tasks have different mini-batch sizes. We find {\em using smaller shot numbers in a task generalizes better especially in the 1-shot meta-test case}. It is also consistent with the notion that difficult (pseudo) tasks (with a smaller support set) facilitate meta-training.

\noindent{\bf The influence of the ``query'' number $Q$.} We also investigate the size of the query set in meta-training. In detail, we fix $K = 1, C = N = 64$ and vary $Q\in\{1,5,10\}$. 
Fig.~\ref{fig:query_bar} illustrates that {\em the number of query instances needs to reach a certain value} to make accurate enough evaluations for few-shot classification in meta-test. But too many query instances will not get further improvements.

\noindent{\bf The influence of the instance number $C$.} We demonstrate the influence of the value $C$ in Fig.~\ref{fig:batch_bar}. We fix $K=1, Q=5$ and vary $C \in \{5,8,16,32,64\}$. 
A larger $C$ indicates more diverse pseudo-classes per episode, and based on our previous discussions, we set $N=C$ when sampling tasks with SES.
Benefited from the proposed SES and SNS, Fig.~\ref{fig:batch_bar} reveals that our UML baseline method is {\em not sensitive to} size of batches, which differs from usual SSL methods~\cite{chen2020simple,chen2020improved}.

\noindent{\bf The influence of different kinds of similarities.} 
We compare various similarities used in the embedding-based classifier in each task, namely, cosine similarity, inner product, (negative) Euclidean distance, and our Semi-Normalized Similarity (SNS). 
We keep the same similarity during meta-training and meta-test while using the default values of $C, N K, Q$ in meta-training. The results of vanilla episodic training, \ie, we only sample one task per episode with SES, is shown in Fig.~\ref{fig:similarity}. Fig. 3 (lower left) in the main paper shows the similarities comparison results when sampling 512 tasks per episode with SES.
We find SNS outperforms other similarities in all cases. For example, it achieves 46.0\% when evaluated on 5-way 1-shot tasks vs. the second best 44.9\%.
Although the superiority gap between SNS and  others becomes smaller when we sampling more tasks with SES, our SNS still gets stable improvements.

\noindent{\bf Summary.} 
The empirical investigations in this section indicate good choices of various factors in meta-training of UML.
In our UML baseline, we sample (pseudo) tasks efficiently with SES and measure similarities between instances with SNS. 
With a fixed size of mini-batch, we increase the ``way'' number $N$ as large as $C$, and optimize over (pseudo) tasks with 1-shot support sets. Since the number of sampled instances $C$ does not make much difference to the performance, we take $C=32$ on account of efficiency.

\begin{table*}[tbp]
	\centering
	\caption{Detailed report of meta-test average few-shot classification accuracy with 95\% confidence interval on different datasets and backbones. Please see Section~\ref{sec:benchmark} for details.}
	\begin{tabular}{lcccc}
		\addlinespace
		\toprule
		{\bf ($N$, $K$)} & {\bf (5,1)} & {\bf (5,5)} & {\bf (5,20)} & {\bf (5,50)} \\
		\midrule
		\multicolumn{5}{c}{{\it Mini}ImageNet with ConvNet} \\
		\midrule
		SimCLR~\cite{chen2020simple} & 42.09 $\pm$ 0.19  & 55.86 $\pm$ 0.17  & 62.17 $\pm$ 0.16  & 63.86 $\pm$ 0.13 \\
		MoCo-v2~\cite{chen2020improved}  & 41.97 $\pm$ 0.19  & 55.00 $\pm$ 0.17  & 60.59 $\pm$ 0.17  & 62.31 $\pm$ 0.14 \\
		MoCHi~\cite{Kalantidis2020Hard} & 40.81 $\pm$ 0.18 & 55.28 $\pm$ 0.16 & 62.89 $\pm$ 0.16 & 65.11 $\pm$ 0.16\\
		\midrule
		CACTUs~\cite{khodadadeh2019unsupervised} {\scriptsize{\bf w/ SES+SNS}} & 44.12 $\pm$ 0.19                        & 58.84 $\pm$ 0.18                        & 66.33 $\pm$ 0.16                        & 68.27 $\pm$ 0.15                        \\
		Baseline (Ours) & 47.43 $\pm$ 0.19 & 64.11 $\pm$ 0.17 & 72.52 $\pm$ 0.15 & 74.72 $\pm$ 0.14 \\
		HMS (Ours) & \textbf{48.12 $\pm$ 0.19} & \textbf{65.33 $\pm$ 0.17} & 73.31 $\pm$ 0.15 & 75.49 $\pm$ 0.13 \\
		TSP-Head (Ours) & 47.35 $\pm$ 0.19 & 65.10 $\pm$ 0.17 &\textbf{ 74.45 $\pm$ 0.14} & \textbf{77.03 $\pm$ 0.13} \\
		
		\midrule
		\multicolumn{5}{c}{{\it Mini}ImageNet with ResNet-12} \\
		\midrule
		SimCLR~\cite{chen2020simple} & 57.75 $\pm$ 0.20 & 72.84 $\pm$ 0.16  & 78.45 $\pm$ 0.13 & 79.75 $\pm$ 0.12 \\
		MoCo-v2~\cite{chen2020improved}  &54.92 $\pm$ 0.20 & 71.18 $\pm$ 0.15 & 77.64 $\pm$ 0.13  & 79.30 $\pm$ 0.11 \\
		MoCHi~\cite{Kalantidis2020Hard} &57.64 $\pm$ 0.20 & 75.53 $\pm$ 0.15& 82.91 $\pm$ 0.12 & 84.76 $\pm$ 0.11 \\
		
				\midrule
		CACTUs~\cite{khodadadeh2019unsupervised} {\scriptsize{\bf w/ SES+SNS}} &57.25  $\pm$    0.22                    & 71.49  $\pm$  0.17                      & 76.27  $\pm$  0.15                      & 77.59  $\pm$  0.14                      \\
		
		Baseline (Ours) &  56.74 $\pm$ 0.20 & 74.05 $\pm$ 0.16 & 81.24 $\pm$ 0.13 & 83.04 $\pm$ 0.12 \\
		HMS (Ours) &\textbf{58.20 $\pm$ 0.20} & 75.77 $\pm$ 0.16 & 82.69 $\pm$ 0.13 & 84.41 $\pm$ 0.12 \\
		TSP-Head (Ours) &56.99 $\pm$ 0.20 & \textbf{75.89 $\pm$ 0.15} & \textbf{83.77 $\pm$ 0.12} & \textbf{85.72 $\pm$ 0.11} \\
		\midrule
		\multicolumn{5}{c}{CIFAR-FS with ResNet-12} \\
		\midrule
		SimCLR~\cite{chen2020simple} &  53.86 $\pm$ 0.19  & 69.19 $\pm$ 0.18  & 75.22 $\pm$ 0.15  & 77.11 $\pm$ 0.14   \\
		MoCo-v2~\cite{chen2020improved}  & 49.73 $\pm$ 0.20  & 64.81 $\pm$ 0.19 & 71.14 $\pm$ 0.15  & 72.70 $\pm$ 0.13  \\
		MoCHi~\cite{Kalantidis2020Hard} & 48.42 $\pm$ 0.22 & 63.91 $\pm$ 0.20 & 71.10 $\pm$ 0.18 & 72.90 $\pm$ 0.17 \\
		
						\midrule
		CACTUs~\cite{khodadadeh2019unsupervised} {\scriptsize{\bf w/ SES+SNS}} &52.53 $\pm$ 0.11   & 73.57  $\pm$  0.19 & 78.48  $\pm$  0.17 & 79.80  $\pm$  0.16 \\
		
		Baseline (Ours) & 53.25 $\pm$ 0.21 & 72.05 $\pm$ 0.18 & 80.03 $\pm$ 0.15 & 82.16 $\pm$ 0.14 \\
		HMS (Ours)&52.20 $\pm$ 0.20 & 72.23 $\pm$ 0.18 & \textbf{82.08 $\pm$ 0.15} & \textbf{84.51 $\pm$ 0.13} \\
		TSP-Head (Ours)&\textbf{54.65 $\pm$ 0.20} & \textbf{73.70 $\pm$ 0.18} & 81.67 $\pm$ 0.15 & 83.86 $\pm$ 0.14 \\
		\midrule
		
		\multicolumn{5}{c}{FC-100 with ResNet-12} \\
		\midrule
		SimCLR~\cite{chen2020simple} &   34.69 $\pm$ 0.16  & 47.07 $\pm$ 0.18  & 54.87 $\pm$ 0.17  & 57.54 $\pm$ 0.16  \\
		MoCo-v2~\cite{chen2020improved}  &	32.86 $\pm$ 0.16 & 44.08 $\pm$ 0.17 & 51.55 $\pm$ 0.17  & 54.15 $\pm$ 0.16  \\
		MoCHi~\cite{Kalantidis2020Hard} &	35.51 $\pm$ 0.17 & 46.95 $\pm$ 0.17 & 54.52 $\pm$ 0.16 & 57.14 $\pm$ 0.16 \\
								\midrule
		CACTUs~\cite{khodadadeh2019unsupervised} {\scriptsize{\bf w/ SES+SNS}} &36.79  $\pm$  0.18 & 48.60  $\pm$  0.18 & 55.52  $\pm$  0.16 & 57.85  $\pm$  0.16 \\
		Baseline (Ours) & 37.31 $\pm$ 0.17 & 51.62 $\pm$ 0.18 & 61.80 $\pm$ 0.17 & 65.54 $\pm$ 0.16 \\
		HMS (Ours)&\textbf{37.88 $\pm$ 0.16} & \textbf{53.68  $\pm$ 0.18} & \textbf{65.14 $\pm$ 0.17} & \textbf{69.15 $\pm$ 0.16} \\
		TSP-Head (Ours)&36.83 $\pm$ 0.16 & 51.78 $\pm$ 0.17 & 62.73 $\pm$ 0.17 & 66.56 $\pm$ 0.16 \\
				\midrule
		\multicolumn{5}{c}{{\it Tiered}ImageNet with ResNet-12} \\
		\midrule
SimCLR~\cite{chen2020simple}          & 57.87 $\pm$ 0.23            & 73.74 $\pm$ 0.19            & 79.71 $\pm$ 0.16            & 81.18 $\pm$ 0.15            \\
MoCo-v2~\cite{chen2020improved}         & 56.63 $\pm$ 0.22            & 74.98 $\pm$ 0.18            & 82.14 $\pm$ 0.15            & 83.98 $\pm$ 0.14            \\
MoCHi~\cite{Kalantidis2020Hard}           & 58.40 $\pm$ 0.22            & 75.49 $\pm$ 0.18            & 82.08 $\pm$ 0.15            & 83.87 $\pm$ 0.14            \\
\midrule
CACTUs~\cite{khodadadeh2019unsupervised} {\scriptsize{\bf w/ SES+SNS}}   & 56.61 $\pm$ 0.23          & 71.22 $\pm$ 0.20          & 76.34 $\pm$ 0.17          & 77.69 $\pm$ 0.17          \\ 
Baseline (Ours) & 56.29 $\pm$ 0.21          & 74.11 $\pm$ 0.18          & 81.64 $\pm$ 0.15          & 83.70 $\pm$ 0.14          \\
HMS (Ours)      & \textbf{58.42 $\pm$ 0.22} & \textbf{75.85 $\pm$ 0.18} & \textbf{82.58 $\pm$ 0.15} & 84.24 $\pm$ 0.14          \\
TSP-Head (Ours) & 56.46 $\pm$ 0.20          & 74.85 $\pm$ 0.16          & 82.56 $\pm$ 0.12          & \textbf{84.52 $\pm$ 0.11} \\
\midrule
\multicolumn{5}{c}{{\it Mini}ImageNet with WRN} \\
		\midrule
SimCLR~\cite{chen2020simple}          & 51.10 $\pm$ 0.20            & 66.52 $\pm$ 0.16            & 72.66 $\pm$ 0.14            & 74.19 $\pm$ 0.13            \\
MoCo-v2~\cite{chen2020improved}         & 50.93  $\pm$  0.20            & 64.59 $\pm$ 0.17            & 72.44 $\pm$ 0.14            & 74.23 $\pm$ 0.14            \\
MoCHi~\cite{Kalantidis2020Hard}           & 53.67  $\pm$  0.20            & 64.24 $\pm$ 0.17            & 71.31 $\pm$ 0.15            & 73.16 $\pm$ 0.14            \\
\midrule
CACTUs~\cite{khodadadeh2019unsupervised} {\scriptsize{\bf w/ SES+SNS}}     & 50.58 $\pm$ 0.21          & 65.10 $\pm$ 0.17          & 70.98 $\pm$ 0.15          & 72.62 $\pm$ 0.14          \\
Baseline (Ours) & 51.47  $\pm$  0.82          & 67.10  $\pm$  0.16          & 74.38  $\pm$  0.14          & 76.37 $\pm$ 0.13          \\
HMS (Ours)      & \textbf{54.13 $\pm$ 0.21} & \textbf{70.01 $\pm$ 0.16} & 75.37 $\pm$ 0.14          & 76.64 $\pm$ 0.13          \\
TSP-Head (Ours) & 51.50 $\pm$ 0.20          & 67.55 $\pm$ 0.17          & \textbf{74.99 $\pm$ 0.14} & \textbf{76.86 $\pm$ 0.13} \\
		\bottomrule
	\end{tabular}
	\label{tab:confidence}
\end{table*}

\section{Benchmark Results}\label{sec:benchmark}
We include the concrete results over various benchmarks in Table~\ref{tab:confidence}.
All methods are evaluated over 10,000 trials. The mean few-shot classification accuracy as well as 95\% confidence interval are reported.

In addition to our UML variants, we also compare with some re-implemented self-supervised learning methods such as SimCLR~\cite{chen2020simple}, MoCo-v2~\cite{chen2020improved}, and MoCHi~\cite{Kalantidis2020Hard}. 
We conclude that our UML methods show great superiority in shallow networks (ConvNet) and higher shots. In these cases, our methods usually surpass self-supervised learning methods by a large margin. 

The results of one representative UML method CACTUs~\cite{khodadadeh2019unsupervised} are also listed, which constructs pseudo-labels by clustering over pre-learned embeddings. We try various ways to obtain the embeddings for clustering, including those used in~\cite{khodadadeh2019unsupervised} and the previous three self-supervised learning methods. The best results among using different pre-learned embeddings are listed as the results of CACTUs in Table~\ref{tab:confidence}. Note that for fair comparisons, we equip CACTUs with our proposed SES and SNS. Our UML methods perform better than CACTUs in all cases.

We also compare various methods based on the Wide ResNet 28-10 backbone (abbreviated as WRN)~\cite{Zagoruyko2016Wide,Rusu2018Meta,YeHZS2018Learning}. The results are shown in the last part of Table~\ref{tab:confidence}, which demonstrates similar phenomena as those using other backbones --- HMS outperforms others on {\em lower shots} tasks while TSP-Head works the best on {\em higher shots} tasks.

\section{Ablation Studies}
In this section, we provide more ablation studies on our UML methods on {\it Mini}ImageNet with ResNet-12 backbone.
\begin{table*}[htbp]
\centering
\caption{Few-shot classification accuracy of both self-supervised learning methods and our UML methods when meta-training with different data augmentation strategies. Methods are evaluated over 10,000 $N$-way $K$-shot tasks on {\it Mini}ImageNet with ResNet-12 backbone.}
\begin{tabular}{cccccc}
\toprule
\textbf{method }         & \textbf{augmentation} &\textbf{ (5,1) }                & \textbf{(5,5) }                & \textbf{(5,20) }               & \textbf{(5,50)}                \\
\midrule
		SimCLR~\cite{chen2020simple}&AMDIM~\cite{Bachman2019Learning} & 57.75 $\pm$ 0.20 & 72.84 $\pm$ 0.16  & 78.45 $\pm$ 0.13 & 79.75 $\pm$ 0.12 \\
		MoCo-v2~\cite{chen2020improved}&AMDIM~\cite{Bachman2019Learning}  &54.92 $\pm$ 0.20 & 71.18 $\pm$ 0.15 & 77.64 $\pm$ 0.13  & 79.30 $\pm$ 0.11 \\
		MoCHi~\cite{Kalantidis2020Hard}&AMDIM~\cite{Bachman2019Learning} &57.64 $\pm$ 0.20 & 75.53 $\pm$ 0.15& 82.91 $\pm$ 0.12 & 84.76 $\pm$ 0.11 \\
		\midrule
		Baseline (Ours)&AMDIM~\cite{Bachman2019Learning} &  56.74 $\pm$ 0.20 & 74.05 $\pm$ 0.16 & 81.24 $\pm$ 0.13 & 83.04 $\pm$ 0.12 \\
		HMS (Ours)&AMDIM~\cite{Bachman2019Learning} &\textbf{58.20 $\pm$ 0.20} & 75.77 $\pm$ 0.16 & 82.69 $\pm$ 0.13 & 84.41 $\pm$ 0.12 \\
		TSP-Head (Ours)&AMDIM~\cite{Bachman2019Learning} &56.99 $\pm$ 0.20 & \textbf{75.89 $\pm$ 0.15} & \textbf{83.77 $\pm$ 0.12} & \textbf{85.72 $\pm$ 0.11} \\ 
		\midrule\midrule
SimCLR~\cite{chen2020simple}          & SimCLR~\cite{chen2020simple}       & 55.21  $\pm$  0.19          & 72.94  $\pm$  0.15          & 80.66  $\pm$  0.12          & 82.59  $\pm$  0.11          \\
MoCo-v2~\cite{chen2020improved}         & SimCLR~\cite{chen2020simple}       & 56.50  $\pm$  0.20          & 73.33  $\pm$  0.16          & 79.41  $\pm$  0.13          & 80.80  $\pm$  0.13          \\
MoCHi~\cite{Kalantidis2020Hard}           & SimCLR~\cite{chen2020simple}       & 54.94  $\pm$  0.20          & 72.68  $\pm$  0.15          & 80.30  $\pm$  0.12          & 82.18  $\pm$  0.11          \\
\midrule
Baseline (Ours) & SimCLR~\cite{chen2020simple}       & 55.33  $\pm$  0.20          & 72.58  $\pm$  0.16          & 79.98  $\pm$  0.13          & 81.87  $\pm$  0.12          \\
HMS (Ours)      & SimCLR~\cite{chen2020simple}       & \textbf{57.21  $\pm$  0.20} & \textbf{74.81  $\pm$  0.16} & 81.26  $\pm$  0.13          & 83.20  $\pm$  0.12          \\
TSP-Head (Ours) & SimCLR~\cite{chen2020simple}       & 55.65  $\pm$  0.20          & 73.18  $\pm$  0.16          & \textbf{81.96  $\pm$  0.13} & \textbf{83.72  $\pm$  0.12} \\ 
\midrule\midrule
SimCLR~\cite{chen2020simple}          & AutoAug~\cite{Cubuk2018Auto}      & 49.01 $\pm$ 0.20            & 65.98 $\pm$ 0.18            & 75.03 $\pm$ 0.15            & 77.64 $\pm$ 0.13            \\
MoCo-v2~\cite{chen2020improved}         & AutoAug~\cite{Cubuk2018Auto}      & 48.37 $\pm$ 0.20            & 64.47 $\pm$ 0.18            & 72.96 $\pm$ 0.16            & 75.56 $\pm$ 0.14            \\
MoCHi~\cite{Kalantidis2020Hard}           & AutoAug~\cite{Cubuk2018Auto}      & 52.69 $\pm$ 0.20            & 70.08 $\pm$ 0.17            & 79.14 $\pm$ 0.14            & 81.40 $\pm$ 0.13            \\
\midrule
Baseline (Ours) & AutoAug~\cite{Cubuk2018Auto}      & 51.49  $\pm$  0.20          & 69.43  $\pm$  0.18          & 77.86  $\pm$  0.15          & 80.08  $\pm$  0.14          \\
HMS (Ours)      & AutoAug~\cite{Cubuk2018Auto}      & 52.91  $\pm$  0.21          & 70.76  $\pm$  0.18          & 78.67  $\pm$  0.15          & 80.80  $\pm$  0.14          \\
TSP-Head (Ours) & AutoAug~\cite{Cubuk2018Auto}      & \textbf{52.93  $\pm$  0.20} & \textbf{70.91  $\pm$  0.17} & \textbf{79.53  $\pm$  0.14} & \textbf{81.76  $\pm$  0.13} \\ 
\midrule\midrule
SimCLR~\cite{chen2020simple}          & RandAug~\cite{Cubuk2020Rand}      & 54.22  $\pm$  0.19          & 72.80  $\pm$  0.15          & 80.59  $\pm$  0.12          & 82.13  $\pm$  0.11          \\
MoCo-v2~\cite{chen2020improved}         & RandAug~\cite{Cubuk2020Rand}      & 55.21  $\pm$  0.20          & 73.07  $\pm$  0.15          & 80.25  $\pm$  0.13          & 81.91  $\pm$  0.12          \\
MoCHi~\cite{Kalantidis2020Hard}           & RandAug~\cite{Cubuk2020Rand}      & 56.27  $\pm$  0.20          & 75.55  $\pm$  0.15          & 81.04  $\pm$  0.13          & 82.44  $\pm$  0.12          \\
\midrule
Baseline (Ours) & RandAug~\cite{Cubuk2020Rand}      & 57.12  $\pm$  0.20          & 75.47  $\pm$  0.16          & 82.13  $\pm$  0.13          & 83.80  $\pm$  0.12          \\
HMS (Ours)      & RandAug~\cite{Cubuk2020Rand}      & \textbf{58.04  $\pm$  0.21} & 74.92  $\pm$  0.16          & 82.21  $\pm$  0.13          & 84.28  $\pm$  0.11          \\
TSP-Head (Ours) & RandAug~\cite{Cubuk2020Rand}      & 56.86  $\pm$  0.20          & \textbf{75.70  $\pm$  0.15} & \textbf{83.38  $\pm$  0.12} & \textbf{85.39  $\pm$  0.11} \\ \bottomrule
\end{tabular}\label{tab:augmentation}
\end{table*}

\subsection{The influence of Data Augmentations}
During meta-training of UML, we generate (pseudo) tasks based on data augmentations. In table 3 in the main paper, we investigate the influence of different data augmentation strategies on various UML methods based on ConvNet. Here we provide the results based on ResNet. We report the few-shot classification performance of both self-supervised learning methods and our UML methods when meta-training with different data augmentation strategies in Table~\ref{tab:augmentation}. 
We find with deeper backbones, RandAug~\cite{Cubuk2020Rand} improves our UML baseline more than AMDIM~\cite{Bachman2019Learning}, but with AMDIM our UML methods get the best results in general. 

\subsection{Comparison with Sampling Strategies}
We compare our SES with ``Hard Task'' sampling method~\cite{Sun2019Meta,Sun2020Meta} (denoted as  and ``HT'' for short). HT constructs hard tasks in meta-training in a two-stage manner and improves supervised FSL. HT first selects hard classes candidates from various tasks and then organizes those hard classes together to get hard tasks.
SES is more efficient since it does not require multiple forward passes per gradient descent step.

We compare SES with the vanilla sampling (with only 1 task per episode) and HT on {\it Mini}ImageNet with both ConvNet and ResNet backbones. 
Detailed $N$-way $K$-shot results are reported in Table~\ref{tab:sampling}. Our SES works better than vanilla sampling and HT consistently. 
The reasons why HT cannot help may come from two factors. First, HT relies on semantic information to construct hard tasks, which becomes difficult in the UML scenario. 
Moreover, the hard (pseudo) class in one task may not still be hard when it is combined with others.

\begin{table*}[t]
	\centering
	\caption{UML comparisons on {\it Mini}ImageNet with ConvNet and ResNet. Methods are evaluated over different $N$-way $K$-shot tasks, and we report the mean classification accuracy over 10,000 trials. For fair comparisons, we use the same similarity, \ie, SES, in the vanilla method (sampling only 1 task) and that with the Hard Task Sampling~\cite{Sun2019Meta,Sun2020Meta} strategy.}
	
	\begin{tabular}{ccccc||cccc}
		\addlinespace
		\toprule
		& \multicolumn{4}{c||}{ConvNet} & \multicolumn{4}{c}{ResNet}\\
		\midrule
		{\bf ($N$, $K$)} & {\bf (5,1)} & {\bf (5,5)} & {\bf (5,20)} & {\bf (5,50)} & {\bf (5,1)} & {\bf (5,5)} & {\bf (5,20)} & {\bf (5,50)} \\
		\midrule
		Vanilla & 43.01 & 57.94  & 65.46  & 67.62 & 48.52  & 65.77& 73.98  & 76.20     \\
		Hard Task~\cite{Sun2019Meta,Sun2020Meta} &  46.37 & 62.88  & 72.28 & 73.86 & 54.74 & 72.61 & 78.96  & 80.93 \\
        \midrule
		SES (Ours) & 47.43  & 64.11  & 72.52  & 74.72 & 56.74  & 74.05  & 81.24  & 83.04  \\
		\bottomrule
	\end{tabular}
	\label{tab:sampling}
\end{table*}

\begin{figure}[t]
	\centering
	\includegraphics[width=0.9\linewidth]{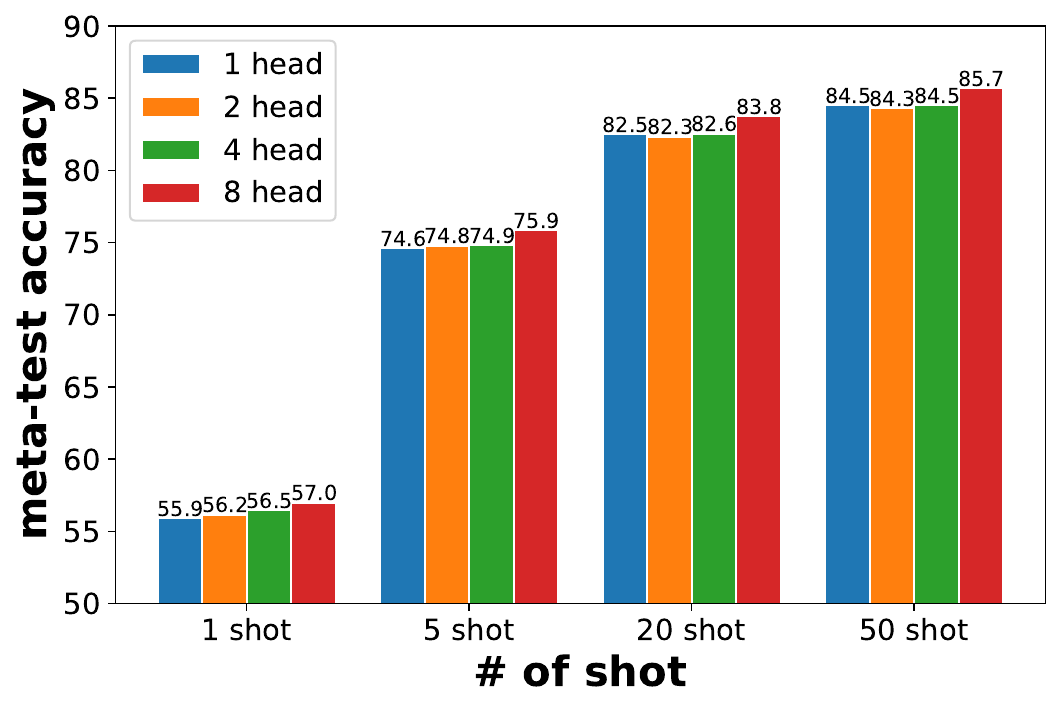}
	\caption{The influence of the number of Transformer heads used in TSP-Head. The number of layers is fixed to 1. Reported results are the averaged 5-way $\{1,5,20,50\}$-shot meta-test accuracy over 10,000 trials on {\it Mini}ImageNet with ResNet backbone. {\it Using more heads in TSP-Head makes the learned embedding generalizable better}.}
	\label{fig:multi-head}
\end{figure}

\begin{figure}[t]
	\centering
	\includegraphics[width=0.9\linewidth]{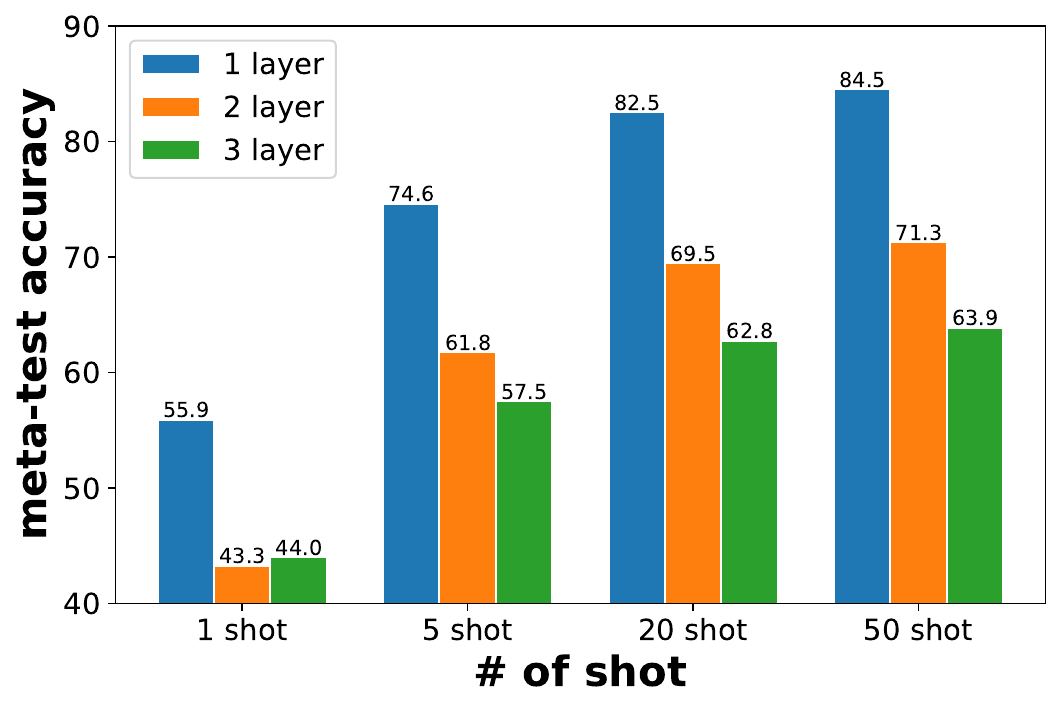}
	\caption{The influence of the number of Transformer layers used in TSP-Head. The number of heads is fixed to 1, and each layer of the Transformer projects the input into vectors with the same dimensionality. Reported results are the averaged 5-way $\{1,5,20,50\}$-shot meta-test accuracy over 10,000 trials on {\it Mini}ImageNet with ResNet backbone. Results show that {\em using one Transformer layer in TSP-Head performs the best}.}
	\label{fig:multi-layer}
\end{figure}

\subsection{Transformer configurations in TSP-head.} As we stated in Section 5.2 in the main paper, the Transformer could be processed multiple times (a.k.a. multi-layer). More than one set of projection matrices could be allocated, and the multiple adapted embeddings could be concatenated followed by a linear projection to dimensionality $d$ (a.k.a. multi-head). We investigate whether such more complicated versions of Transformer will help TSP-Head. The comparison results of different setups are shown in Fig.~\ref{fig:multi-head} and Fig.~\ref{fig:multi-layer}, respectively.

Based on the results in Fig.~\ref{fig:multi-head}, we find {\em increasing the number of Transformer heads does help learn better embeddings}. In contrast, the results in Fig.~\ref{fig:multi-layer} indicate {\em using more transformer layers leads to the performance drop}. Therefore, we use an 8-head 1-layer version of Transformer in TSP-Head. 
It is notable that the configuration of the Transformer is different from a recent supervised usage of Transformer for few-shot classification FEAT~\cite{YeHZS2018Learning}. In FEAT, the Transformer is used in both meta-training and meta-test. In other words, the post-adapted embeddings with Transformer are used in the embedding-based few-shot classifier in meta-test. Experiments in~\cite{YeHZS2018Learning} show only one head and one layer works the best with FEAT.
In TSP-Head, we use Transformer only in unsupervised meta-training. The results in Fig.~\ref{fig:multi-head} indicate that a stronger Transformer with more heads increases the generalization ability of the pre-adapted embedding.
\begin{table}[tbp]
	\centering
	\caption{Few-shot classification accuracy of FEAT and MetaOptNet on {\it Mini}ImageNet with ResNet-12 backbone, which utilize adapted embeddings during both meta-training and meta-test. 
	``FEAT'' applies the Transformer head over the support set during meta-test. We also include the results using the learned Transformer head in both support and query sets for evaluation as TSP-Head works in meta-training, denoted as ``w/ head''.
	Methods are evaluated over 10,000 $N$-way $K$-shot tasks.
	}
	\begin{tabular}{lcccc}
		\addlinespace
		\toprule
{\bf ($N$, $K$)} & \textbf{(5,1)}                & \textbf{(5,5)}                & \textbf{(5,20)}               & \textbf{(5,50)}               \\
\midrule
\multicolumn{5}{l}{\it Meta-test with post-adapted embeddings} \\ 
MetaOptNet~\cite{Lee2019Meta}      & 56.71 & 68.11 & 74.65 & 79.02 \\
FEAT~\cite{YeHZS2018Learning}            & 56.67 & 73.77 & 81.09 & 82.97 \\
w/ head                & 56.90          & 75.87          & 83.59         & 85.69         \\
\midrule
\multicolumn{5}{l}{\it Meta-test with pre-adapted embeddings} \\
TSP-Head        & \textbf{56.99} & \textbf{75.89} & \textbf{83.77} & \textbf{85.72} \\
		\bottomrule
	\end{tabular}
	\label{tab:feat_opt}
\end{table}
\begin{table}[tbp]
	\centering
	\caption{Mean classification accuracy between task-agnostic projection head / task-specific projection head on different $N$-way $K$-shot tasks over 10,000 trials on {\it Mini}ImageNet. Task-specific projection head outperforms task-agnostic ones in almost all setups, especially those with higher shots. Besides, our Transformer-based TSP-head performs better than the FiLM-based TSP-head in all cases.}
	
	\begin{tabular}{ccccc}
		\addlinespace
		\toprule
		{\bf ($N$, $K$)} & {\bf (5,1)} & {\bf (5,5)} & {\bf (5,20)} & {\bf (5,50)}\\
		\midrule
		\multicolumn{5}{l}{\it Task-agnostic projection head} \\
		Projection Head~\cite{chen2020simple} & 53.32 & 70.87 & 78.38 & 80.32\\
		\midrule
\multicolumn{5}{l}{\it Task-specific projection head} \\
		DeepSets~\cite{Zaheer17Deep}+FiLM~\cite{Perez2018FILM} & 51.71 & 73.28 & 82.15 &84.48 \\
		TSP-Head (ours) & \bf 56.99  & \bf 75.89 & \bf 83.77 & \bf 85.72 \\
		\bottomrule
	\end{tabular}
	\label{tab:sup-projection-head}
\end{table}
\subsection{Maintaining TSP-Head in Meta-Test or Not?}
In our TSP-Head, we meta-train the post-adapted embedding $\psi = \mathbf{T} \circ \phi$ while use the pre-adapted vanilla embedding $\phi$ during meta-test. We expect that $\psi$ decomposes the special and general properties of tasks into the transformation $\mathbf{T}$ and $\phi$, respectively, so that $\phi$ could be more generalizable.

We verify that TSP-Head achieves better results than the UML baseline especially evaluated with {\em higher} shots, but a natural question is whether maintaining the transformation $\mathbf{T}$ in meta-test could get further improvement. 
When using TSP-Head, we transform all embeddings in both support and query sets during meta-training, however, since we cannot collect all query set instances in advance during meta-test, the learned TSP-Head can only be applied to the support set of novel classes.
We compare TSP-Head with FEAT~\cite{YeHZS2018Learning}, which constructs the embedding-based classifier with the adapted task-specific embeddings during meta-test. 
FEAT applies the Transformer head over the support set during meta-test. We also include the results using the learned Transformer head in both support and query sets during evaluation for reference, which is denoted as ``w/ head''.
``w/ head'' works in a transductive manner, the same as what TSP-Head works in meta-training (see also Table 8 in the main paper).
The results are listed in Table~\ref{tab:feat_opt}, which shows using the pre-adapted task-agnostic embedding $\phi$ outperforms using the task-specific embedding $\psi$. 

Since using the post-adapted embeddings in meta-test equals applying embedding adaptation methods based on the sampled (pseudo) tasks in both meta-training and meta-test, we also compare with another embedding adaptation method MetaOptNet~\cite{Lee2019Meta}. 
MetaOptNet adapts embeddings with a task-specific convex optimization.
From the results in Table~\ref{tab:feat_opt}, MetaOptNet also degrades a lot during meta-test. 

Therefore, results in Table~\ref{tab:feat_opt} reveal that TSP-Head makes the pre-adapted embedding $\phi$ more generalizable than the post-adapted one. One possible reason could be the inconsistency of task generation between meta-training and meta-test --- the embedding optimizes over (pseudo) tasks during meta-training, which pulls different views of an instance together and pushes any two instances away. In meta-test, however, the learned embedding is asked to discern few-shot classification tasks with semantically similar or dissimilar classes (based on ground-truth class labels).

\subsection{Configurations of DeepSets and FiLM}
We study other ways to implement the transformation $\mathbf{T}$ in the TSP-Head besides Transformer~\cite{Vaswani2017Attention}.
Inspired from~\cite{Oreshkin2018TADAM}, we consider FiLM~\cite{Perez2018FILM} layer as another effective choice of the task-specific transformation. 
FiLM generates task-specific scale and bias to make the transformation task-dependent. 
Following~\cite{YeHZS2018Learning}, we use DeepSets~\cite{Zaheer17Deep} to capture the characteristic of a task. To be specific, for instances in the union of support and query sets, $\x \in \mathcal{S} \cup \mathcal{Q}$, its embedding $\phi(x)$ is transformed with the following manner
\begin{align}
    (\boldsymbol{\gamma} ,\boldsymbol{\beta}) &= h\left(\sum_{\x' \in (\mathcal{S},\mathcal{Q})} f\left(\phi\left(\x'\right)\right) \right)\;,\label{eq:deepsets}\\
    \psi(\x) &= \boldsymbol{\gamma} \odot \phi(\x) + \boldsymbol{\beta}\;.
\end{align}
where we implement $h$ and $f$ as two-layer MLPs. $\boldsymbol{\gamma}$ and $\boldsymbol{\beta}$ has the same dimension as $\phi(x)$. $\odot$ is element-wise product of two vectors. The summation in Eq.~\ref{eq:deepsets} makes the transformation permutation invariant~\cite{Zaheer17Deep,YeHZS2018Learning}.

Similar to our Transformer-based projection head, FiLM-based projection head adapts $\phi(\x)$ with $\psi(\x)$ and the post-adapted $\psi(\x)$ is used to calculate the meta-training objective. Once finishing meta-training, the pre-adapted embedding $\phi(\x)$ is used in meta-test. Results in Table~\ref{tab:sup-projection-head} indicate that the Transformer implementation of TSP-Head works better than the FiLM implementation. Besides, task-specific projection head outperforms task-agnostic ones in almost all the setup, especially higher shots.

\begin{table}[t]
	\centering
	\caption{The change of performance of HMS when meta-trained with different configurations of SES. We increase the number of task per-episode in SES from 1 to 512. Methods are evaluated over 10,000 $N$-way $K$-shot tasks on {\it Mini}ImageNet with ResNet-12 backbone  and average accuracy is reported.}
\begin{tabular}{@{}ccccc@{}}
\toprule
 \textbf{\# of Tasks}   & \textbf{(5,1)}        & \textbf{(5,5)}        & \textbf{(5,20)}       & \textbf{(5,50)}       \\ \midrule
 1   & 50.98  & 69.30  & 78.73  & 81.10  \\
 8   & 54.83  & 73.02  & 81.02  & 83.16  \\
 64  & 55.97  & 74.16  & 82.11  & 83.99  \\
 256 & 58.20  & 75.77  & 82.69  & 84.41  \\
 512 & \textbf{58.51} & \textbf{76.27} & \textbf{83.28} & \textbf{84.97} \\ \bottomrule
\end{tabular}
\label{tb:task_hms}
\end{table}
\subsection{The Influence of Task Number in HMS}
We report the results of HMS based on the usage of SES in the main paper. In this subsection, we investigate how the number of sampled (pseudo) tasks per episode influences the performance of HMS. 
In Table~\ref{tb:task_hms}, we gradually increase the task number in SES from 1 to 512. Results show that HMS benefits from the increased number of tasks, \eg, from 50.98\% to 58.51\% evaluated on 5-way 1-shot tasks. 
Therefore, instead of choosing hard negatives from a large memory bank~\cite{Kalantidis2020Hard}, our HMS takes advantage of the re-sampled tasks per episode and makes the meta-learned embeddings discriminative.





%

%

\end{document}